\documentclass{article}

\usepackage{amsmath,amsfonts,amssymb,amsthm,bm}
\usepackage{enumitem}
\usepackage{microtype}
\usepackage{graphbox}
\usepackage{graphicx}
\usepackage{caption}
\usepackage{subcaption}
\usepackage{booktabs} 
\usepackage{amsfonts}       
\usepackage{nicefrac}       

\newcommand{\E}{\mathbb{E}}

\newcommand{\bs}{\boldsymbol}

\usepackage{xcolor}
\definecolor{C0}{HTML}{1f77b4}
\definecolor{C1}{HTML}{ff7f0e}
\definecolor{C2}{HTML}{2ca02c}
\definecolor{C3}{HTML}{d62728}
\definecolor{C4}{HTML}{9467bd}
\definecolor{C5}{HTML}{8c564b}
\definecolor{C6}{HTML}{e377c2}
\definecolor{C7}{HTML}{7f7f7f}
\definecolor{C8}{HTML}{bcbd22}
\definecolor{C9}{HTML}{17becf}
\usepackage{wrapfig}
\usepackage{siunitx}
\usepackage{array, booktabs, makecell}
\usepackage{multirow}

\usepackage{color}
\usepackage{xcolor}
\usepackage{pifont}
\newcommand{\cmark}{\textcolor{green!80!black}{\ding{51}}}
\newcommand{\xmark}{\textcolor{red}{\ding{55}}}

\newif\ifshow
\showtrue
\showfalse

\ifshow
\newcommand{\ari}[1]{\textcolor{blue}{[\textbf{Ari:} #1]}}
\newcommand{\stephane}[1]{\textcolor{brown}{[\textbf{Stephane:} #1]}}
\newcommand{\levent}[1]{\textcolor{olive}{[\textbf{Levent}: #1]}}
\newcommand{\matthew}[1]{\textcolor{magenta}{[\textbf{Matthew}: #1]}}
\newcommand{\giulio}[1]{\textcolor{cyan}{[\textbf{Giulio}: #1]}}
\newcommand{\hugo}[1]{\textcolor{violet}{[\textbf{Hugo}: #1]}}
\newcommand{\todo}[1]{\textcolor{red}{[\textbf{TODO}: #1]}}

\else
\newcommand{\ari}[1]{}
\newcommand{\stephane}[1]{}
\newcommand{\levent}[1]{}
\newcommand{\hugo}[1]{}
\newcommand{\matthew}[1]{}
\newcommand{\giulio}[1]{}
\newcommand{\todo}[1]{}

\fi

\usepackage{hyperref}

\usepackage[accepted]{icml2021}

\icmltitlerunning{ConViT: Improving Vision Transformers with Soft Convolutional Inductive Biases}

\begin{document}

\twocolumn[
\icmltitle{ConViT: Improving Vision Transformers\\ with Soft Convolutional Inductive Biases}



\icmlsetsymbol{equal}{*}

\begin{icmlauthorlist}
\icmlauthor{Stéphane d'Ascoli}{ens,fair}
\icmlauthor{Hugo Touvron}{fair}
\icmlauthor{Matthew L. Leavitt}{fair}
\icmlauthor{Ari S. Morcos}{fair}
\icmlauthor{Giulio Biroli}{ens,fair}
\icmlauthor{Levent Sagun}{fair}
\end{icmlauthorlist}

\icmlaffiliation{ens}{Department of Physics, Ecole Normale Sup\'erieure, Paris, France}
\icmlaffiliation{fair}{Facebook AI Research, Paris, France}

\icmlcorrespondingauthor{Stéphane d'Ascoli}{stephane.dascoli@ens.fr}

\icmlkeywords{Convolutional networks, Vision Transformers, Inductive biases, Initialization}

\vskip 0.3in
]



\printAffiliationsAndNotice{}  

\begin{abstract}
Convolutional architectures have proven extremely successful for vision tasks. Their hard inductive biases enable sample-efficient learning, but come at the cost of a potentially lower performance ceiling. Vision Transformers (ViTs) rely on more flexible self-attention layers, and have recently outperformed CNNs for image classification. However, they require costly pre-training on large external datasets or distillation from pre-trained convolutional networks. In this paper, we ask the following question: is it possible to combine the strengths of these two architectures while avoiding their respective limitations? To this end, we introduce \emph{gated positional self-attention} (GPSA), a form of positional self-attention which can be equipped with a ``soft" convolutional inductive bias. We initialize the GPSA layers to mimic the locality of convolutional layers, then give each attention head the freedom to escape locality by adjusting a \emph{gating parameter} regulating the attention paid to position versus content information. The resulting convolutional-like ViT architecture, \textit{ConViT}, outperforms the DeiT~\cite{touvron2020training} on ImageNet, while offering a much improved sample efficiency.
We further investigate the role of locality in learning by first quantifying how it is encouraged in vanilla self-attention layers, then analyzing how it is escaped in GPSA layers. We conclude by presenting various ablations to better understand the success of the ConViT. Our code and models are released publicly at \url{https://github.com/facebookresearch/convit}.
\end{abstract}

\section{Introduction}

The success of deep learning over the last decade has largely been fueled by models with strong inductive biases, allowing efficient training across domains~\cite{mitchell1980need,goodfellow_deep_2016}. The use of Convolutional Neural Networks (CNNs)~\cite{lecun1998gradient,lecun1989backpropagation}, which have become ubiquitous in computer vision since the success of AlexNet in 2012~\cite{krizhevsky2017imagenet}, epitomizes this trend. Inductive biases are hard-coded into the architectural structure of CNNs in the form of two strong constraints on the weights: locality and weight sharing. By encouraging translation equivariance (without pooling layers) and translation invariance (with pooling layers)~\cite{scherer_evaluation_2010,schmidhuber_deep_2015,goodfellow_deep_2016}, the convolutional inductive bias makes models more sample-efficient and parameter-efficient~\cite{simoncelli2001natural,ruderman1994statistics}. Similarly, for sequence-based tasks, recurrent networks with hard-coded memory cells have been shown to simplify the learning of long-range dependencies (LSTMs) and outperform vanilla recurrent neural networks in a variety of settings~\cite{gers1999learning,sundermeyer_lstm_2012,greff_lstm_2017}.

However, the rise of models based purely on attention in recent years calls into question the necessity of hard-coded inductive biases. First introduced as an add-on to recurrent neural networks for Sequence-to-Sequence models~\cite{bahdanau2014neural}, attention has led to a breakthrough in Natural Language Processing through the emergence of Transformer models, which rely solely on a particular kind of attention: Self-Attention (SA) \cite{vaswani2017attention}. The strong performance of these models when pre-trained on large datasets has quickly led to Transformer-based approaches becoming the default choice over recurrent models like LSTMs~\cite{devlin2018bert}.

In vision tasks, the locality of CNNs impairs the ability to capture long-range dependencies, whereas attention does not suffer from this limitation. \citet{chen20182} and ~\citet{bello2019attention} leveraged this complementarity by augmenting convolutional layers with attention. More recently, ~\citet{ramachandran2019stand} ran a series of experiments replacing some or all convolutional layers in ResNets with attention, and found the best performing models used convolutions in early layers and attention in later layers. The Vision Transformer (ViT), introduced by ~\citet{dosovitskiy2020image}, entirely dispenses with the convolutional inductive bias by performing SA across embeddings of patches of pixels. The ViT is able to match or exceed the performance of CNNs but requires pre-training on vast amounts of data. More recently, the Data-efficient Vision Transformer (DeiT)~\cite{touvron2020training} was able to reach similar performances without any pre-training on supplementary data, instead relying on Knowledge Distillation~\cite{hinton2015distilling} from a convolutional teacher.

\begin{figure}[t]
    \centering
    \includegraphics[width=\columnwidth]{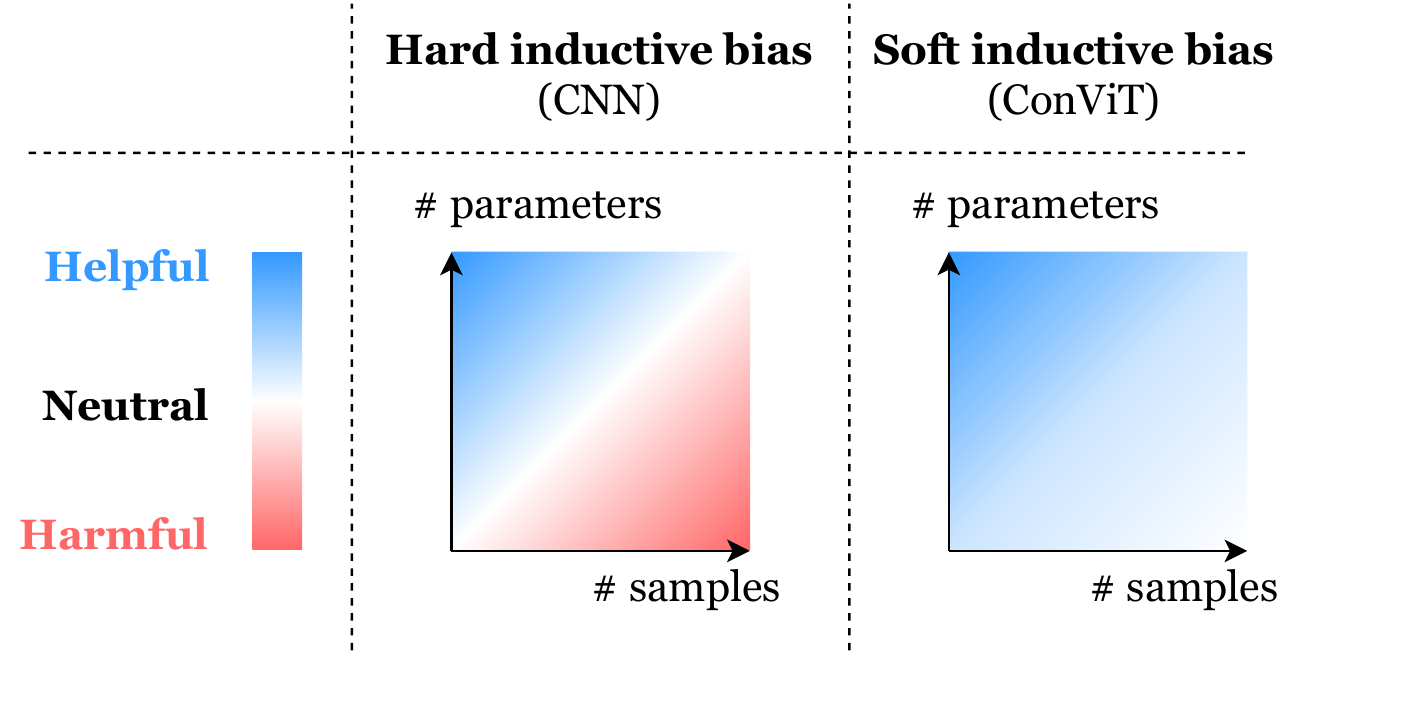}
    \caption{\textbf{Soft inductive biases can help models learn without being restrictive.} Hard inductive biases, such as the architectural constraints of CNNs, can greatly improve the sample-efficiency of learning, but can become constraining when the size of the dataset is not an issue. The soft inductive biases introduced by the ConViT avoid this limitation by vanishing away when not required. 
    }
    \label{fig:phase_space}
\end{figure}

\paragraph{Soft inductive biases}

The recent success of the ViT demonstrates that while convolutional constraints can enable strongly sample-efficient training in the small-data regime, they can also become limiting as the dataset size is not an issue. In data-plentiful regimes, hard inductive biases can be overly restrictive and \textit{learning} the most appropriate inductive bias can prove more effective. The practitioner is therefore confronted with a dilemma between using a convolutional model, which has a high performance floor but a potentially lower performance ceiling due to the hard inductive biases, or a self-attention based model, which has a lower floor but a higher ceiling. This dilemma leads to the following question: can one get the best of both worlds, and obtain the benefits of the convolutional inductive biases without suffering from its limitations (see Fig.~\ref{fig:phase_space})?

In this direction, one successful approach is the combination of the two architectures in ``hybrid" models. These models, which interleave or combine convolutional and self-attention layers, have fueled successful results in a variety of tasks ~\cite{carion2020end,hu2018relation,ramachandran2019stand,chen2020uniter,locatello2020object,sun2019videobert,srinivas20201bottleneck,wu_visual_2020}. Another approach is that of Knowledge Distillation~\cite{hinton2015distilling}, which has recently been applied to transfer the inductive bias of a convolutional teacher to a student transformer~\cite{touvron2020training}. While these two methods offer an interesting compromise, they forcefully induce convolutional inductive biases into the Transformers, potentially affecting the Transformer with their limitations. 

\begin{figure}[t]
    \centering
    \begin{subfigure}[b]{.49\columnwidth}
    \includegraphics[width=\linewidth]{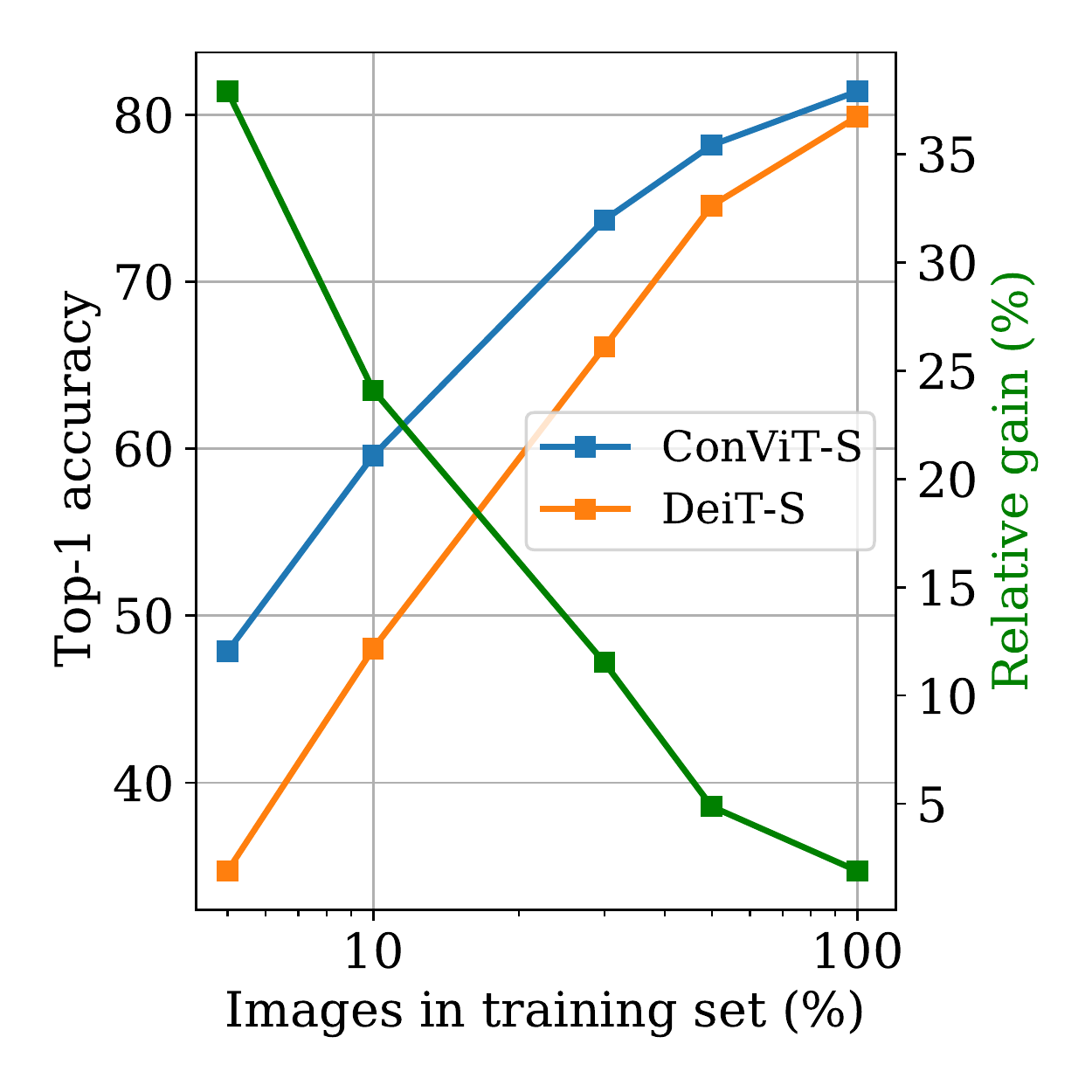}    
    \caption{Sample efficiency}
    \end{subfigure}
    \begin{subfigure}[b]{.49\columnwidth}
    \includegraphics[width=\linewidth]{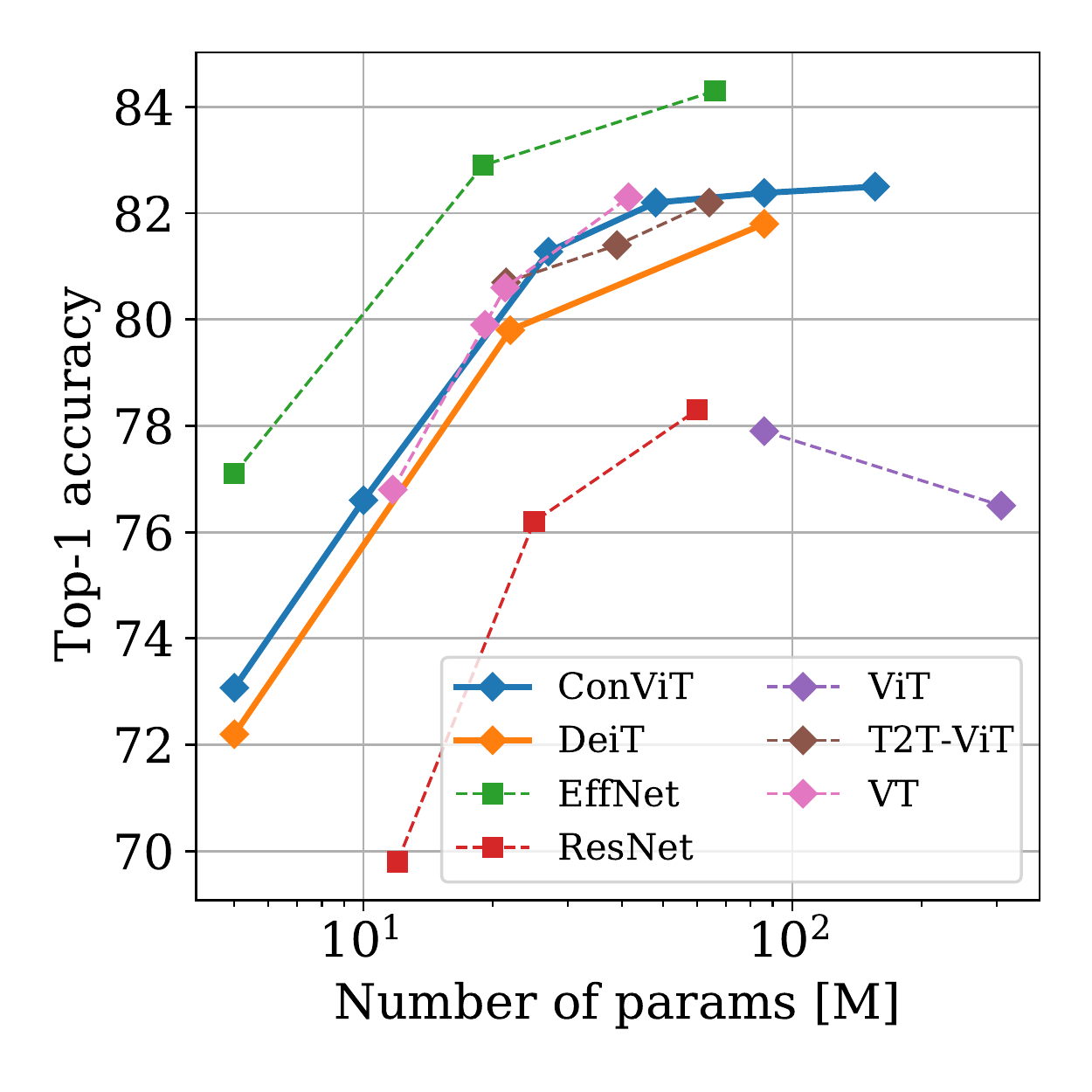}    
    \caption{Parameter efficiency}
    \end{subfigure}
    \caption{\textbf{The ConViT outperforms the DeiT both in sample and parameter efficiency}. \textit{Left:} we compare the sample efficiency of our ConViT-S (see Tab.~\ref{tab:statistics}) with that of the DeiT-S by training them on restricted portions of ImageNet-1k, where we only keep a certain fraction of the images of each class. Both models are trained with the hyperparameters reported in~\cite{touvron2020training}. We display the the relative improvement of the ConViT over the DeiT in green. \textit{Right:} we compare the top-1 accuracies of our ConViT models with those of other ViTs (diamonds) and CNNs (squares) on ImageNet-1k. The performance of other models on ImageNet are taken from~\cite{touvron2020training,he2016deep,tan2019efficientnet,wu_visual_2020,yuan2021tokens}.}
    \label{fig:sota}
\end{figure}

\paragraph{Contribution}
In this paper, we take a new step towards bridging the gap between CNNs and Transformers, by presenting a new method to ``softly" introduce a convolutional inductive bias into the ViT. The idea is to let each SA layer decide whether to behave as a convolutional layer or not, depending on the context. We make the following contributions:
\begin{enumerate}[noitemsep,labelindent=0pt]
    \item We present a new form of SA layer, named \emph{gated positional self-attention} (GPSA), which one can initialize as a convolutional layer. Each attention head then has the freedom to recover expressivity by adjusting a \emph{gating parameter}.
    \item We then perform experiments based on the DeiT~\cite{touvron2020training}, with a certain number of SA layers replaced by GPSA layers. The resulting Convolutional Vision Transformer (ConViT) outperforms the DeiT while boasting a much improved sample-efficiency (Fig.~\ref{fig:sota}). 
    \item We analyze quantitatively how local attention is naturally encouraged in vanilla ViTs, then investigate the inner workings of the ConViT and perform ablations to investigate how it benefits from the convolution initialization.
\end{enumerate}

Overall, our work demonstrates the effectiveness of "soft" inductive biases, especially in the low-data regime where the learning model is highly underspecified (see Fig.~\ref{fig:phase_space}), and motivates the exploration of further methods to induce them. 

\giulio{This last sentence coudl be expanded a bit and here we can refer to Fig.1} \levent{indeed, DeiT is a bit of a specific choice and would not be directly familiar for many people, we could add a few sentences justifying the choice of DeiT and describing it's core contribution in a sentence (which we already expand a bit in the `related work' session @hugo)}

\paragraph{Related work}
\matthew{I feel like there's also an angle of our work being relevant to the broader goal of making models more sample-efficient. Perhaps could start general and move towards the two more specific examples you describe below.}

Our work is motivated by combining the recent success of pure Transformer models ~\cite{dosovitskiy2020image} with the formalized relationship between SA and convolution. Indeed, \citet{cordonnier2019relationship} showed that a SA layer with $N_h$ heads can express a convolution of kernel size $\sqrt N_h$, if each head focuses on one of the pixels in the kernel patch. By investigating the qualitative aspect of attention maps of models trained on CIFAR-10, it is shown that SA layers with relative positional encodings naturally converge towards convolutional-like configurations, suggesting that some degree of convolutional inductive bias is desirable.

Conversely, the restrictiveness of hard locality constraints has been proven by~\citet{elsayed2020revisiting}. A breadth of approaches have been taken to imbue CNN architectures with nonlocality ~\cite{hu_gather-excite_2018,hu_squeeze-and-excitation_2018,wang_non-local_2018,wu_visual_2020}. Another line of research is to induce a convolutional inductive bias is different architectures. For example, \citet{neyshabur2020towards} uses a regularization method to encourage fully-connected networks (FCNs) to learn convolutions from scratch throughout training. 

Most related to our approach, \citet{d2019finding} explored a method to initialize FCNs networks as CNNs. This enables the resulting FCN to reach much higher performance than achievable with standard initialization. Moreover, if the FCN is initialized from a partially trained CNN, the recovered degrees of freedom allow the FCN to outperform the CNN it stems from. This method relates more generally to ``warm start" approaches such as those used in spiked tensor models \cite{anandkumar2016homotopy}, where a smart initialization, containing prior information on the problem, is used to ease the learning task.

\paragraph{Reproducibility}
We provide an open-source implementation of our method as well as pretrained models at the following address: \url{https://github.com/facebookresearch/convit}.

\section{Background}

We begin by introducing the basics of SA layers, and show how positional attention can allow SA layers to express convolutional layers.

\paragraph{Multi-head self-attention} \label{sec:mhsa}

The attention mechanism is based on a trainable associative memory with (key, query) vector pairs. A sequence of $L_1$ ``query" embeddings $\boldsymbol Q\in \mathbb{R}^{L_1\times D_{h}}$ is matched against another sequence of $L_2$ ``key" embeddings $\boldsymbol K\in\mathbb{R}^{L_2\times D_{h}}$ using inner products. The result is an attention matrix whose entry $(ij)$ quantifies how semantically ``relevant" $\boldsymbol Q_i$ is to $\boldsymbol K_j$:
\begin{equation}
    \boldsymbol A = \operatorname{softmax}\left(\frac{\boldsymbol Q \boldsymbol K^\top}{\sqrt {D_{h}}}\right) \in \mathbb{R
}^{L_1\times L_2},
    \label{eq:attention}
\end{equation}
where $\left(\operatorname{softmax}\left[\bs X\right]\right)_{ij} = \bs e^{\bs X_{ij}} / \sum_k e^{\bs X_{ik}}$.

Self-attention is a special case of attention where a sequence is matched to itself, to extract the semantic dependencies between its parts. In the ViT, the queries and keys are linear projections of the embeddings of $16\times 16$ pixel patches $\boldsymbol X\in\mathbb{R}^{L\times D_\mathrm{emb}}$. Hence, we have $\boldsymbol Q = \boldsymbol W_{qry} \boldsymbol X$ and $\boldsymbol K = \boldsymbol W_{key} \boldsymbol X$, where $\boldsymbol{W}_{key}, \boldsymbol{W}_{qry}\in\mathbb R^{D_\mathrm{emb}\times D_{h}}$. 

Multi-head SA layers use several self-attention heads in parallel to allow the learning of different kinds of interdependencies. They take as input a sequence of $L$ embeddings of dimension $D_\mathrm{emb} = N_h D_h$, and output a sequence of $L$ embeddings of the same dimension through the following mechanism:
\begin{align}
    \operatorname{MSA}(\boldsymbol{X}):=\underset{h \in\left[N_{h}\right]}{\operatorname{concat}}\left[\text{SA}_{h}(\boldsymbol{X})\right] \boldsymbol{W}_{out}+\boldsymbol{b}_{out},
\end{align}
where $\boldsymbol{W}_{out}\in\mathbb R^{D_\mathrm{emb} \times D_\mathrm{emb}}, b_{out}\in\mathbb R^{D_\mathrm{emb}}$. Each self-attention head $h$ performs the following operation:  
\begin{align}
    \text{SA}_h(\boldsymbol{X}) := \boldsymbol{A}^h \boldsymbol{X} \boldsymbol{W}_{val}^h,
\end{align}
where $\boldsymbol W_{val}^h \in R^{D_\mathrm{emb} \times D_h}$ is the \emph{value} matrix.

However, in the vanilla form of Eq.~\ref{eq:attention}, SA layers are position-agnostic: they do not know how the patches are located according to each other. To incorporate positional information, there are several options. One is to add some positional information to the input at embedding time, before propagating it through the SA layers: ~\cite{dosovitskiy2020image} use this approach in their ViT. Another possibility is to replace the vanilla SA with positional self-attention (PSA), using encodings $\boldsymbol r_{ij}$ of the relative position of patches $i$ and $j$~\cite{ramachandran2019stand}:
\begin{align}
    \boldsymbol{A}^h_{ij}:=\operatorname{softmax}\left(\boldsymbol Q^h_i \boldsymbol K^{h\top}_j+\boldsymbol{v}_{pos}^{h\top} \boldsymbol{r}_{ij}\right)
\label{eq:local-attention}
\end{align}
Each attention head uses a trainable embedding $\boldsymbol{v}_{pos}^h \in \mathbb R^{D_\mathrm{pos}}$, and the relative positional encodings $\boldsymbol{r}_{ij}\in \mathbb R^{D_\mathrm{pos}}$ only depend on the distance between pixels $i$ and $j$,  denoted denoted by a two-dimensional vector $\boldsymbol \delta_{ij}$. 

\begin{figure}[tb]
    \centering
    \hspace{-1.5em}
    \begin{subfigure}[b]{.32\columnwidth}
    \includegraphics[width=\linewidth]{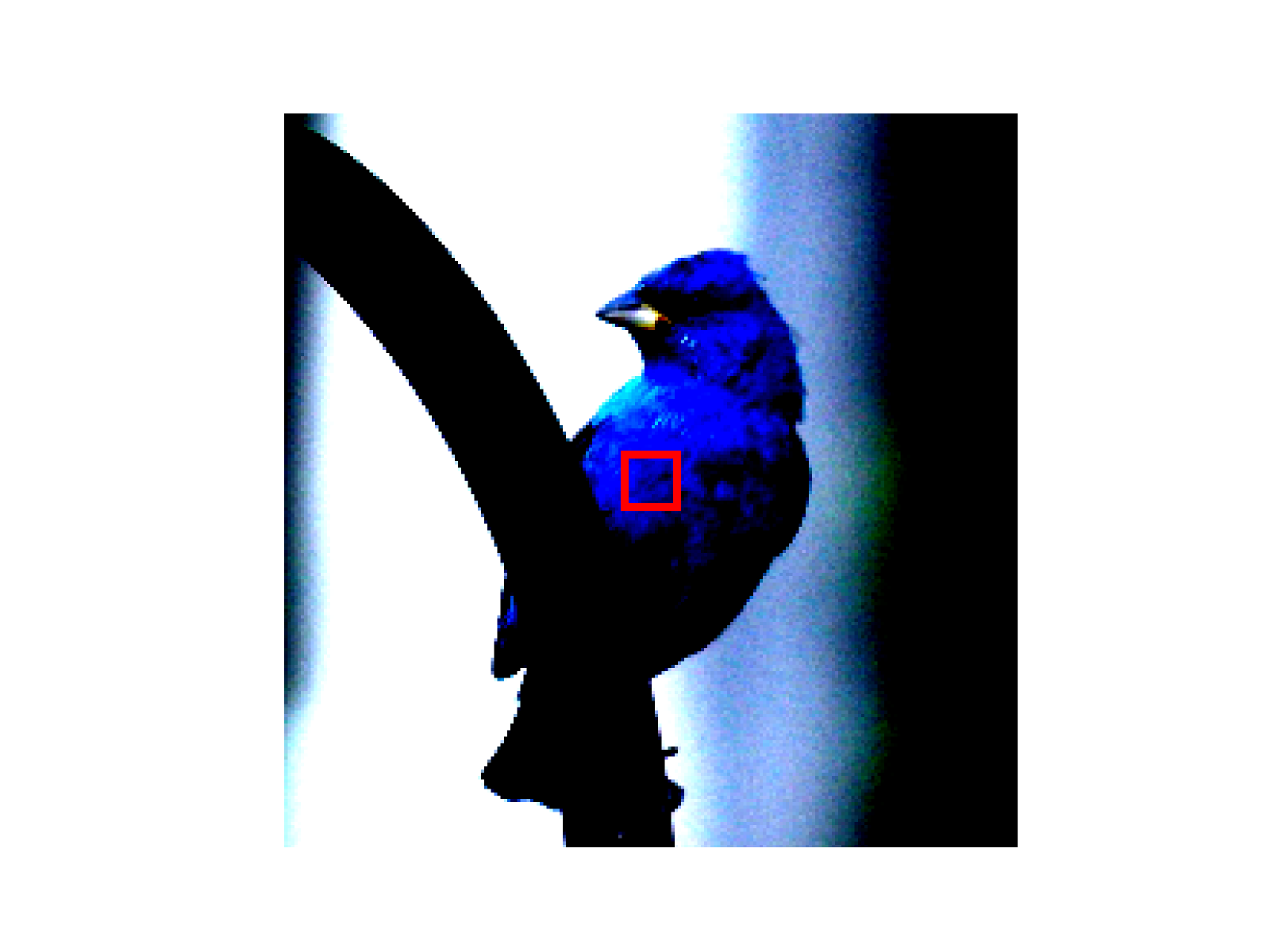}
    \caption{Input}
    \end{subfigure}
    \hspace{-1.1em}
    \begin{subfigure}[b]{.74\columnwidth}
    \includegraphics[width=\linewidth]{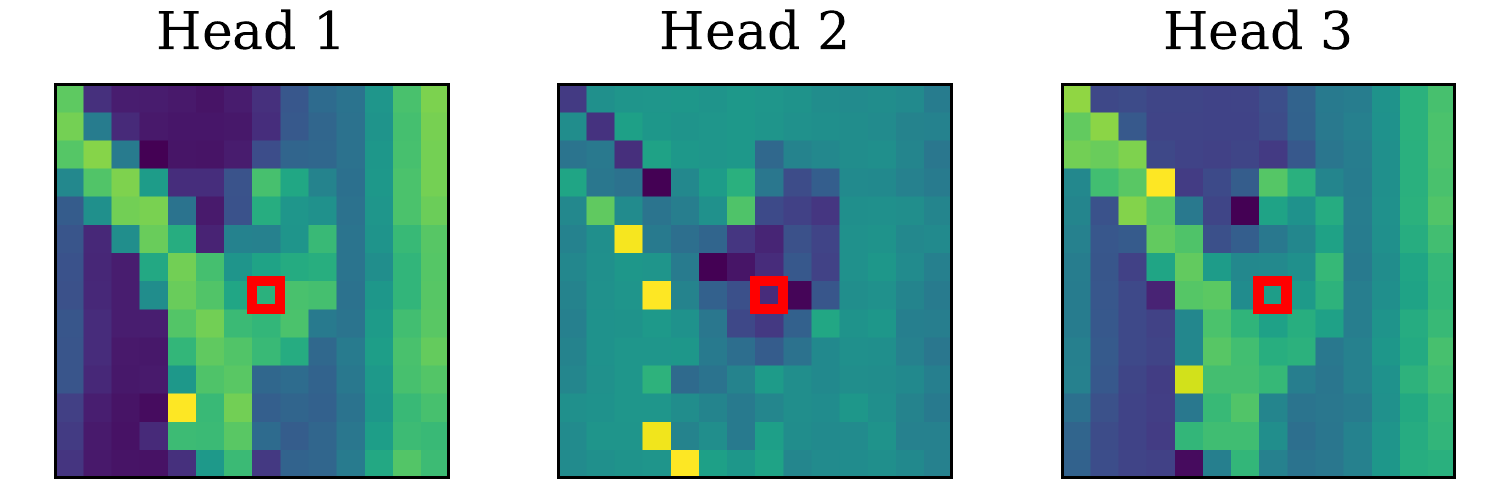}
    \caption{Standard initialization}
    \end{subfigure}
    \begin{subfigure}[b]{\columnwidth}
    \includegraphics[width=\linewidth]{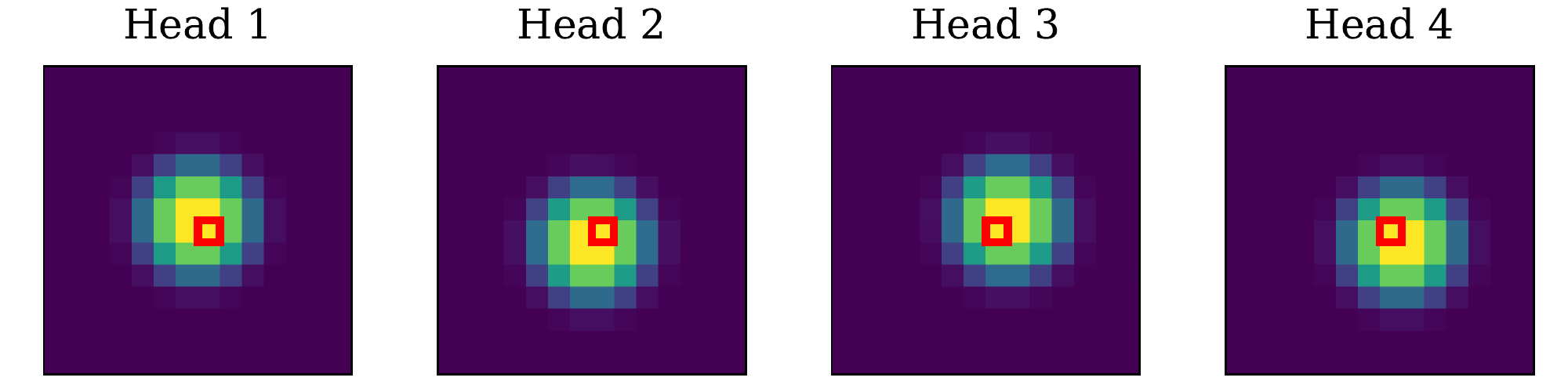}
    \caption{Convolutional initialization, strength $\alpha=0.5$}
    \end{subfigure}   
    \begin{subfigure}[b]{\columnwidth}
    \includegraphics[width=\linewidth]{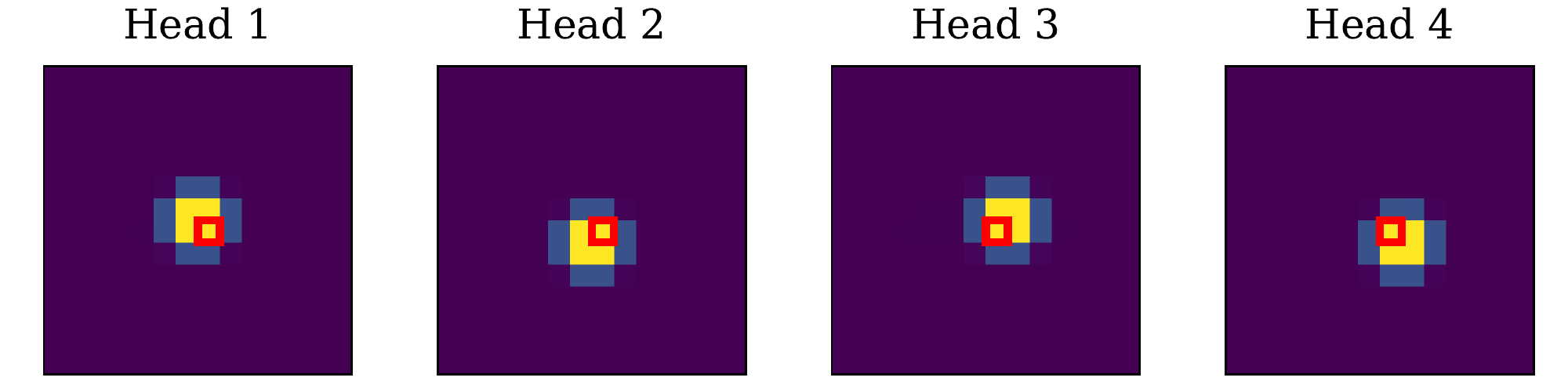}
    \caption{Convolutional initialization, strength $\alpha=2$}
    \end{subfigure}
    \caption{\textbf{Positional self-attention layers can be initialized as convolutional layers.} (a): Input image from ImageNet, where the query patch is highlighted by a red box. (b),(c),(d): attention maps of an untrained SA layer (b) and those of a PSA layer using the convolutional-like initialization scheme of Eq.~\ref{eq:local-init} with two different values of the locality strength parameter, $\alpha$ (c, d). Note that the shapes of the image can easily be distinguished in (b), but not in (c) or (d), when the attention is purely positional.}
    \label{fig:maps-init}
\end{figure}

\paragraph{Self-attention as a generalized convolution}

\citet{cordonnier2019relationship} show that a multi-head PSA layer with $N_h$ heads and learnable relative positional encodings~(Eq. \ref{eq:local-attention}) of dimension $D_\mathrm{pos}\geq 3$ can express any convolutional layer of filter size $\sqrt N_h \times \sqrt N_h$, by setting the following:
\begin{align}
    \begin{cases}
    &\boldsymbol{v}_{pos}^{h}:=-\alpha^{h}\left(1,-2 \Delta_{1}^{h},-2 \Delta_{2}^{h}, 0,\ldots 0\right)\\ &\boldsymbol{r}_{\boldsymbol{\delta}}:=\left(\|\boldsymbol{\delta}\|^{2}, \delta_{1}, \delta_{2},0,\ldots 0\right)\\
    &\boldsymbol{W}_{q r y}=\boldsymbol{W}_{k e y}:=\mathbf{0}, \quad \boldsymbol{W}_{val}:=\boldsymbol{I}\\
    \end{cases}
    \label{eq:local-init}
\end{align}

In the above, 
\begin{itemize}
    \item The \emph{center of attention} $\boldsymbol \Delta^h\in\mathbb{R}^2$ is the position to which head $h$ pays most attention to, relative to the query patch. For example, in Fig.~\ref{fig:maps-init}(c), the four heads correspond, from left to right, to $\boldsymbol \Delta^1 = (-1,1), \boldsymbol \Delta^2 = (-1,-1), \boldsymbol \Delta^3 = (1,1), \boldsymbol \Delta^4 = (1,-1)$.
    \item The \emph{locality strength} $\alpha^h>0$ determines how focused the attention is around its center $\boldsymbol \Delta^h$ (it can also by understood as the ``temperature'' of the softmax in Eq.~\ref{eq:attention}). When $\alpha^h$ is large, the attention is focused only on the patch(es) located at $\boldsymbol \Delta^h$, as in Fig.~\ref{fig:maps-init}(d); when $\alpha^h$ is small, the attention is spread out into a larger area, as in Fig.~\ref{fig:maps-init}(c). 
\end{itemize} 

Thus, the PSA layer can achieve a strictly convolutional attention map by setting the centers of attention $\boldsymbol \Delta^h$ to each of the possible positional offsets of a $\sqrt{N_h}\times \sqrt{N_h}$ convolutional kernel, and sending the locality strengths $\alpha^h$ to some large value. 

\section{Approach}

\begin{figure}[tb]
    \centering
    \includegraphics[width=\columnwidth]{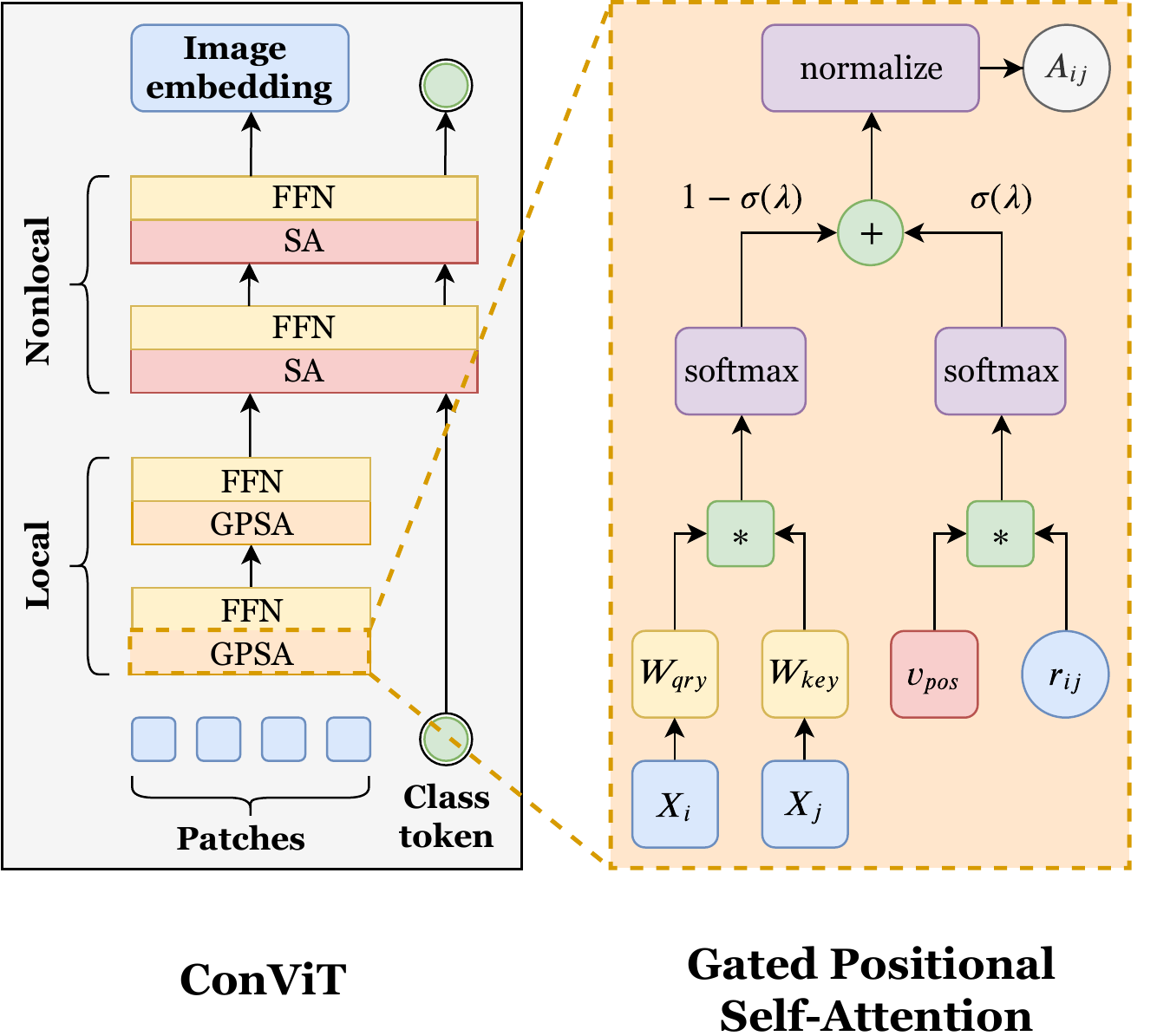}
    \caption{\textbf{Architecture of the ConViT.} The ConViT (left) is a version of the ViT in which some of the self-attention (SA) layers are replaced with gated positional self-attention layers (GPSA; right). Because GPSA layers involve positional information, the class token is concatenated with hidden representation after the last GPSA layer. In this paper, we typically take 10 GPSA layers followed by 2 vanilla SA layers. FFN: feedforward network (2 linear layers separated by a GeLU activation); $W_{qry}$: query weights; $W_{key}$: key weights; $v_{pos}$: attention center and span embeddings (learned); $r_{qk}$: relative position encodings (fixed); $\lambda$: gating parameter (learned); $\sigma$: sigmoid function.}
    \label{fig:architecture}
\end{figure}

Building on the insight of~\cite{cordonnier2019relationship}, we introduce the ConVit, a variant of the ViT~\cite{dosovitskiy2020image} obtained by replacing some of the SA layers by a new type of layer which we call \emph{gated positional self-attention} (GPSA) layers. The core idea is to enforce the ``informed" convolutional configuration of Eqs.~\ref{eq:local-init} in the GPSA layers at initialization, then let them decide whether to stay convolutional or not. However, the standard parameterization of PSA layers (Eq.~\ref{eq:local-attention}) suffers from two limitations, which lead us two introduce two modifications.

\paragraph{Adaptive attention span} The first caveat in PSA is the vast number of trainable parameters involved, since the number of relative positional encodings $\boldsymbol r_\delta$ is quadratic in the number of patches. This led some authors to restrict the attention to a subset of patches around the query patch~\cite{ramachandran2019stand}, at the cost of losing long-range information. 

To avoid this, we leave the relative positional encodings $\boldsymbol r_\delta$ fixed, and train only the embeddings $\boldsymbol v^h_{pos}$ which determine the center and span of the attention heads; this approach relates to the \emph{adaptive attention span} introduced in~\citet{sukhbaatar2019adaptive} for Language Transformers. The initial values of $\boldsymbol r_\delta$ and $\boldsymbol v_{pos}^h$ are given by Eq.~\ref{eq:local-init}, where we take $D_\mathrm{pos}=3$ to get rid of the useless zero components. Thanks to $D_\mathrm{pos}\ll D_h$, the number of parameters involved in the positional attention is negligible compared to the number of parameters involved in the content attention. This makes sense, as content interactions are inherently much simpler to model than positional interactions.

\paragraph{Positional gating} 
The second issue with standard PSA is the fact that the content and positional terms in Eq.~\ref{eq:local-attention} are potentially of different magnitudes, in which case the softmax will ignore the smallest of the two. In particular, the convolutional initialization scheme discussed above involves highly concentrated attention scores, i.e. high-magnitude values in the softmax. In practice, we observed that using a convolutional initialization scheme on vanilla PSA layers gives a boost in early epochs, but degrades late-time performance as the attention mechanism lazily ignores the content information (see SM.~\ref{app:without-gpsa}). 

To avoid this, GPSA layers sum the content and positional terms \emph{after} the softmax, with their relative importances governed by a learnable \emph{gating} parameter $\lambda_h$ (one for each attention head). Finally, we normalize the resulting sum of matrices (whose terms are positive) to ensure that the resulting attention scores define a probability distribution. The resulting GPSA layer is therefore parametrized as follows (see also Fig.~\ref{fig:architecture}):
\begin{align}
    \text{GPSA}_h(\boldsymbol{X}) :=& \operatorname{normalize}\left[\boldsymbol{A}^h\right] \boldsymbol{X} \boldsymbol{W}_{val}^h\\
    \boldsymbol{A}^h_{ij}:=&\left(1-\sigma(\lambda_h)\right) \operatorname{softmax}\left(\boldsymbol Q^h_i \boldsymbol K^{h\top}_j\right) \notag\\
    &+ \sigma(\lambda_h) \operatorname{softmax}\left(\boldsymbol{v}_{pos}^{h\top} \boldsymbol{r}_{ij}\right),
    \label{eq:gating-param}
\end{align}
where $\left(\operatorname{normalize}\left[\bs A\right]\right)_{ij} = \bs A_{ij} / \sum_k \bs A_{ik}$ and $\sigma:x\mapsto \nicefrac{1}{\left(1+e^{-x}\right)}$ is the sigmoid function. By setting the gating parameter $\lambda_h$ to a large positive value at initialization, one has $\sigma(\lambda_h)\simeq 1$ : the GPSA bases its attention purely on position, dispensing with the need of setting $\boldsymbol W_{qry}$ and $\boldsymbol W_{key}$ to zero as in Eq.~\ref{eq:local-init}. However, to avoid the ConViT staying stuck at $\lambda_h\gg 1$, we initialize $\lambda_h = 1$ for all layers and all heads.

\begin{table*}[tb]
    \centering
    \begin{tabular}{c|c|c|c|c|c|c|c|c}
    \toprule
    Name & Model & $N_h$ & $D_\mathrm{emb}$ & Size & Flops & Speed & Top-1 & Top-5\\
    \midrule
    \multirow{2}{*}{Ti} & DeiT & 3 & 192 & 6M & 1G & 1442 & 72.2 & -\\
       & ConViT       & 4 & 192 & 6M & 1G & 734 & \textbf{73.1} & \textbf{91.7}\\
    \hline
    \multirow{2}{*}{Ti+} & DeiT   & 4 & 256 & 10M & 2G & 1036 & 75.9 & 93.2 \\
    & ConViT & 4 & 256 & 10M & 2G & 625 & \textbf{76.7} & \textbf{93.6}\\
    \hline
    \multirow{2}{*}{S} & DeiT & 6 & 384 & 22M & 4.3G & 587  & 79.8 & -\\
    & ConViT       & 9 & 432 & 27M & 5.4G & 305 & \textbf{81.3} & \textbf{95.7}\\
    \hline
    \multirow{2}{*}{S+} & DeiT   & 9 & 576 & 48M & 10G & 480 & 79.0 & 94.4\\
    & ConViT & 9 & 576 & 48M & 10G & 382 & \textbf{82.2} & \textbf{95.9}\\
    \hline
    \multirow{2}{*}{B} & DeiT & 12 & 768 & 86M & 17G & 187 & 81.8 & -\\
    & ConViT       & 16 & 768 & 86M & 17G & 141 & \textbf{82.4} & \textbf{95.9}\\
    \hline
    \multirow{2}{*}{B+} & DeiT   & 16 & 1024 & 152M & 30G & 114 & 77.5 & 93.5\\
    & ConViT & 16 & 1024 & 152M & 30G & 96 & \textbf{82.5} & \textbf{95.9}\\
    \bottomrule
    \end{tabular}
    \caption{\textbf{Performance of the models considered, trained from scratch on ImageNet.} Speed is the number of images processed per second on a Nvidia Quadro GP100 GPU at batch size 128. Top-1 accuracy is measured on ImageNet-1k test set without distillation (see SM.~\ref{app:distillation} for distillation). The results for DeiT-Ti, DeiT-S and DeiT-B are reported from~\cite{touvron2020training}.
    }
    \label{tab:statistics}
\end{table*}

\begin{table}[tb]
    \footnotesize
    \centering
    \begin{tabular}{c|c|c|c|c|c|c}
    \toprule
    \textbf{Train} & \multicolumn{3}{c|}{\textbf{Top-1}} & \multicolumn{3}{c}{\textbf{Top-5}}\\
    \textbf{size} & DeiT & ConViT & Gap & DeiT & ConViT & Gap\\
    \midrule
    5\% & 34.8 & \textbf{47.8} & 37\% & 57.8 & \textbf{70.7} & 22\%\\
    10\% & 48.0 & \textbf{59.6} & 24\% & 71.5 & \textbf{80.3} & 12\%\\
    30\% & 66.1 & \textbf{73.7} & 12\% & 86.0 & \textbf{90.7} & 5\%\\
    50\% & 74.6 & \textbf{78.2} & 5\% & 91.8 & \textbf{93.8} & 2\%\\
    100\%& 79.9 & \textbf{81.4} & 2\% & 95.0 & \textbf{95.8} & 1\%\\
    \bottomrule
    \end{tabular}
    \caption{\textbf{The convolutional inductive bias strongly improves sample efficiency.} We compare the top-1 and top-5 accuracy of our ConViT-S with that of the DeiT-S, both trained using the original hyperparameters of the DeiT~\cite{touvron2020training}, as well as the relative improvement of the ConViT over the DeiT.  Both models are trained on a subsampled version of ImageNet-1k, where we only keep a variable fraction (leftmost column) of the images of each class for training.}
    \label{tab:sample-efficiency}
\end{table}

\paragraph{Architectural details}

The ViT slices input images of size $224$ into $16\times 16$ non-overlapping patches of $14\times 14$ pixels and embeds them into vectors of dimension $D_\mathrm{emb} = 64 N_h$ using a convolutional stem. It then propagates the patches through 12 blocks which keep their dimensionality constant. Each block consists in a SA layer followed by a 2-layer Feed-Forward Network (FFN) with GeLU activation, both equipped with residual connections. The ConViT is simply a ViT where the first 10 blocks replace the SA layers by GPSA layers with a convolutional initialization.

Similar to language Transformers like BERT~\cite{devlin2018bert}, the ViT uses an extra ``class token", appended to the sequence of patches to predict the class of the input. Since this class token does not carry any positional information, the SA layers of the ViT do not use positional attention: the positional information is instead injected to each patch before the first layer, by adding a learnable positional embedding of dimension $D_\mathrm{emb}$. As GPSA layers involve positional attention, they are not well suited for the class token approach. We solve this problem by appending the class token to the patches after the last GPSA layer, similarly to what is done in~\cite{touvron2021going} (see Fig.~\ref{fig:architecture})\footnote{We also experimented incorporating the class token as an extra patch of the image to which all heads pay attention to at initialization, but results were worse than concatenating the class token after the GPSA layers (not shown).}. 

For fairness, and since they are computationally cheap, we keep the absolute positional embeddings of the ViT active in the ConViT. However, as shown in SM.~\ref{app:masking}, the ConViT relies much less on them, since the GPSA layers already use relative positional encodings. Hence, the absolute positional embeddings could easily be removed, dispensing with the need to interpolate the embeddings when changing the input resolution (the relative positional encodings simply need to be resampled according to Eq.~\ref{eq:local-init}, as performed automatically in our open-source implementation).

\paragraph{Training details}

We based our ConVit on the DeiT~\cite{touvron2020training}, a hyperparameter-optimized version of the ViT which has been open-sourced\footnote{ \url{https://github.com/facebookresearch/deit}}. Thanks to its ability to achieve competitive results without using any external data, the DeiT both an excellent baseline and relatively easy to train: the largest model (DeiT-B) only requires a few days of training on 8 GPUs. 

To mimic $2\times 2$, $3\times 3$ and $4\times 4$ convolutional filters, we consider three different ConViT models with 4, 9 and 16 attention heads (see Tab.~\ref{tab:statistics}). Their number of heads are slightly larger than the DeiT-Ti, ConViT-S and ConViT-B of~\citet{touvron2020training}, which respectively use 3, 6 and 12 attention heads. To obtain models of similar sizes, we use two methods of comparison.
\begin{itemize}
    \item To establish a direct comparison with~\citet{touvron2020training}, we lower the embedding dimension of the ConViTs to $D_\mathrm{emb}/N_h = 48$ instead of $64$ used for the DeiTs. Importantly, \textit{we leave all hyperparameters (scheduling, data-augmentation, regularization) presented in~\cite{touvron2020training} unchanged} in order to achieve a fair comparison. The resulting models are named ConViT-Ti, ConViT-S and ConViT-B.
    \item We also trained DeiTs and ConViTs using the same number of heads and $D_\mathrm{emb}/N_h = 64$, to ensure that the improvement due to ConViT is not simply due to the larger number of heads~\cite{touvron2021going}. This leads to slightly larger models denoted with a ``+'' in Tab.~\ref{tab:statistics}. To maintain stable training while fitting these models on 8 GPUs, we lowered the learning rate from $0.0005$ to $0.0004$ and the batch size from $1024$ to $512$. These minimal hyperparameter changes lead the DeiT-B+ to perform less well than the DeiT-S+, which is not the case for the ConViT, suggesting a higher stability to hyperparameter changes.
\end{itemize}

\paragraph{Performance of the ConViT}

In Tab.~\ref{tab:statistics}, we display the top-1 accuracy achieved by these models evaluated on the ImageNet test set after 300 epochs of training, alongside their number of parameters, number of flops and throughput. Each ConViT outperforms its DeiT of same size and same number of flops by a margin. Importantly, although the positional self-attention does slow down the throughput of the ConViTs, they also outperform the DeiTs at equal throughput. For example, The ConViT-S+ reaches a top-1 of $82.2\%$, outperforming the original DeiT-B with less parameters and higher throughput. Without any tuning, the ConViT also reaches high performance on CIFAR100, see SM.~\ref{app:performance} where we also report learning curves.

Note that our ConViT is compatible with the distillation methods introduced in~\citet{touvron2020training} at no extra cost. As shown in~SM. \ref{app:distillation}, hard distillation improves performance, enabling the hard-distilled ConViT-S+ to reach $82.9\%$ top-1 accuracy, on the same footing as the hard-distilled DeiT-B with half the number of parameters. However, while distillation requires an additional forward pass through a pre-trained CNN at each step of training, ConViT has no such requirement, providing similar benefits to distillation without additonal computational requirements. 

\paragraph{Sample efficiency of the ConViT}

In Tab.~\ref{tab:sample-efficiency}, we investigate the sample-efficiency of the ConViT in a systematic way, by subsampling each class of the ImageNet-1k dataset by a fraction $f=\{0.05,0.1,0.3,0.5,1\}$ while multiplying the number of epochs by $1/f$ so that the total number images presented to the model remains constant. As one might expect, the top-1 accuracy of both the DeiT-S and its ConViT-S counterpart drops as $f$ decreases. However, the ConViT suffers much less: while training on only 10\% of the data, the ConVit reaches 59.5\% top-1 accuracy, compared to 46.5\% for its DeiT counterpart.

This result can be directly compared to~\cite{zhai2019s4l}, which after testing several thousand convolutional models reaches a top-1 accuracy of 56.4\%; the ConViT is therefore highly competitive in terms of sample efficiency. These findings confirm our hypothesis that the convolutional inductive bias is most helpful on small datasets, as depicted in Fig.~\ref{fig:phase_space}.

\section{Investigating the role of locality}
\label{sec:experimental}

In this section, we demonstrate that locality is naturally encouraged in standard SA layers, and examine how the ConViT benefits from locality being imposed at initialization.

\paragraph{SA layers are pulled towards locality}

We begin by investigating whether the hypothesis that PSA layers are naturally encouraged to become ``local" over the course of training~\cite{cordonnier2019relationship} holds for the vanilla SA layers used in ViTs, which do not benefit from positional attention. To quantify this, we define a measure of ``nonlocality" by summing, for each query patch $i$, the distances $\Vert \boldsymbol \delta_{ij}\Vert $ to all the key patches $j$ weighted by their attention score $\boldsymbol A_{ij}$. We average the number obtained over the query patch to obtain the nonlocality metric of head $h$, which can then be averaged over the attention heads to obtain the nonlocality of the whole layer $\ell$:
\begin{align}
    D_{loc}^{\ell,h} &:= \frac{1}{L} \sum_{ij} \boldsymbol A^{h,\ell}_{ij} \Vert \boldsymbol \delta_{ij}\Vert, \notag\\
    D_{loc}^{\ell} &:= \frac{1}{N_h} \sum_h D_{loc}^{\ell,h}
    \label{eq:nonlocality}
\end{align}
Intuitively, $D_{loc}$ is the number of patches between the center of attention and the query patch: the further the attention heads look from the query patch, the higher the nonlocality. 

\begin{figure}[t]
    \centering
    \includegraphics[width=\columnwidth]{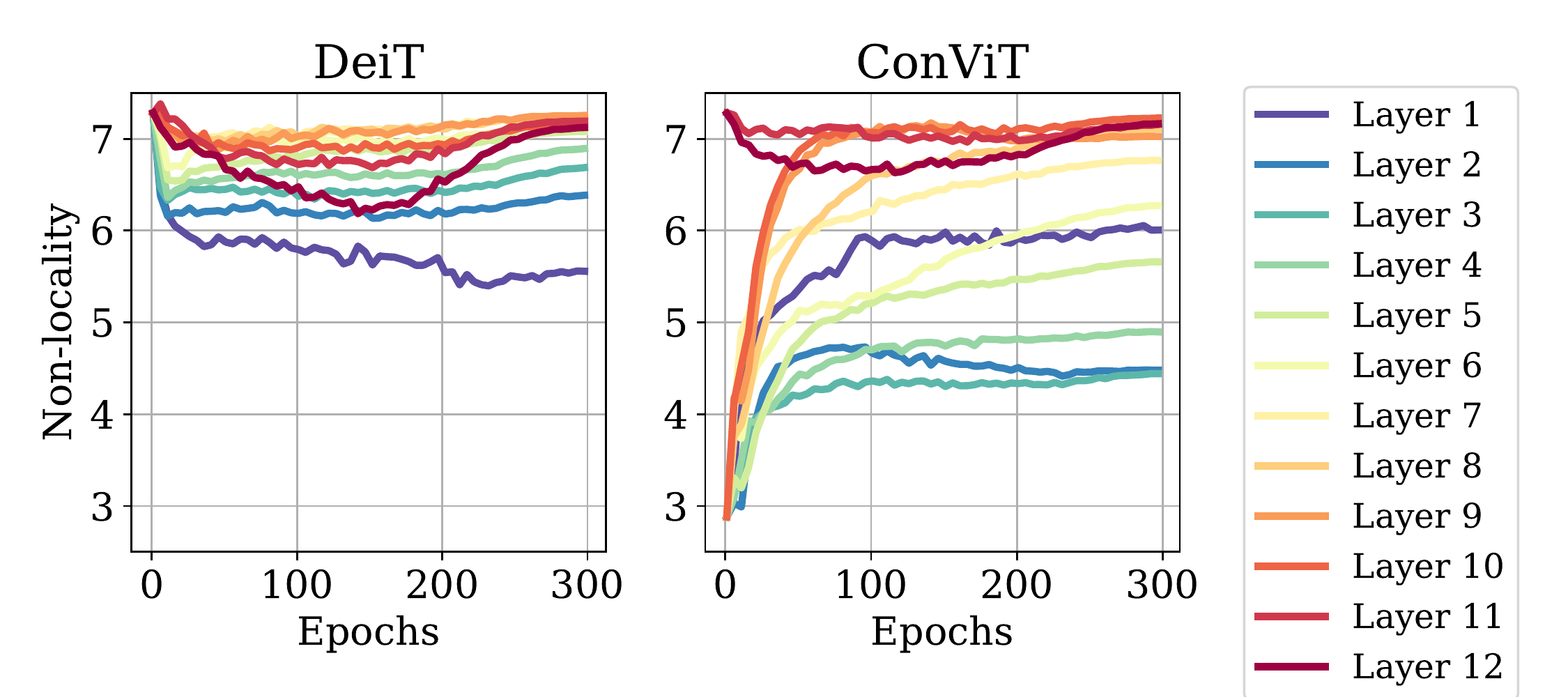}
    \caption{\textbf{SA layers try to become local, GPSA layers escape locality.} We plot the nonlocality metric defined in Eq.~\ref{eq:nonlocality}, averaged over a batch of 1024 images: the higher, the further the attention heads look from the query pixel. We trained the DeiT-S and ConViT-S for 300 epochs on ImageNet. Similar results for DeiT-Ti/ConViT-Ti and DeiT-B/ConViT-B are shown in SM.~\ref{app:nonlocality}.}
    \label{fig:nonlocality}
\end{figure}

In Fig.~\ref{fig:nonlocality} (left panel), we show how the nonlocality metric evolves during training across the 12 layers of a DeiT-S trained for 300 epochs on ImageNet. During the first few epochs, the nonlocality falls from its initial value in all layers, confirming that the DeiT becomes more ``convolutional". During the later stages of training, the nonlocality metric stays low for lower layers, and gradually climbs back up for upper layers, revealing that the latter capture long range dependencies, as observed for language Transformers~\cite{sukhbaatar2019adaptive}. 

These observations are particularly clear when examining the attention maps (Fig.~\ref{fig:attn-nonlocal} of the SM), and point to the beneficial effect of locality in lower layers. In Fig.~\ref{fig:nonlocality-distillation} of the SM., we also show that the nonlocality metric is lower when training with distillation from a convolutional network as in~\citet{touvron2020training},  suggesting that the locality of the teacher is partly transferred to the student~\cite{abnar2020transferring}.

\begin{figure}[t]
    \centering
    \includegraphics[width=\linewidth]{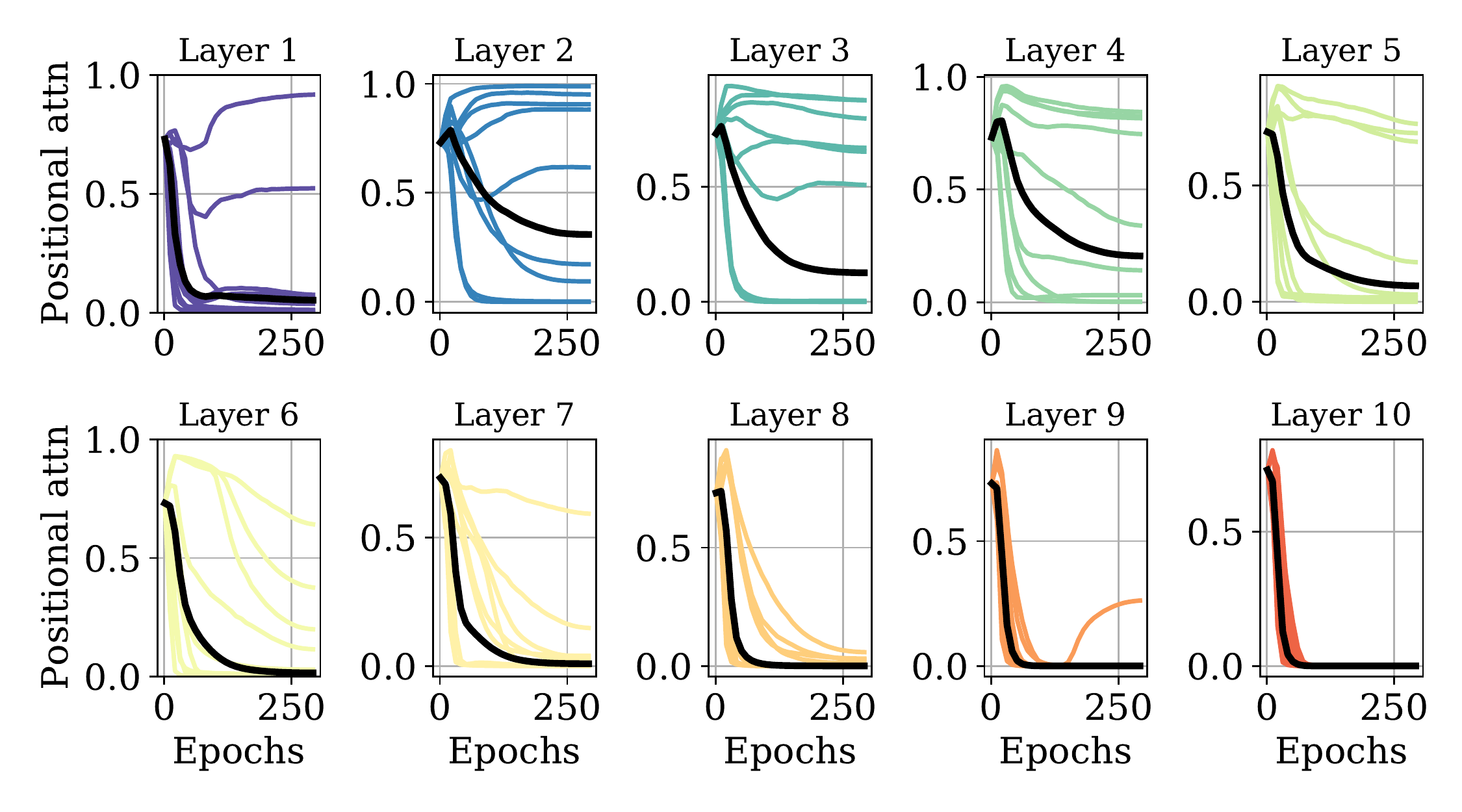}
    \caption{\textbf{The gating parameters reveal the inner workings of the ConViT.} For each layer, the colored lines (one for each of the 9 attention heads) quantify how much attention head $h$ pays to positional information versus content, i.e. the value of $\sigma(\lambda_h)$, see Eq.~\ref{eq:gating-param}. The black line represents the value averaged over all heads. We trained the ConViT-S for 300 epochs on ImageNet. Similar results for ConViT-Ti and ConViT-B are shown in SM~\ref{app:nonlocality}.}
    \label{fig:gating}
\end{figure}

\paragraph{GPSA layers escape locality}

In the ConViT, strong locality is imposed at the beginning of training in the GPSA layers thanks to the convolutional initialization. In Fig.~\ref{fig:nonlocality} (right panel), we see that this local configuration is escaped throughout training, as the nonlocality metric grows in all the GPSA layers. However, the nonlocality at the end of training is lower than that reached by the DeiT, showing that some information about the initialization is preserved throughout training. Interestingly, the final nonlocality does not increase monotonically throughout the layers as for the DeiT. The first layer and the final layers strongly escape locality, whereas the intermediate layers (particularly the second layer) stay more local. 

To gain more understanding, we examine the dynamics of the gating parameters in Fig.~\ref{fig:gating}. We see that in all layers, the average gating parameter $\E_h\sigma(\lambda_h)$ (in black), which reflects the average amount of attention paid to positional information versus content, decreases throughout training. This quantity reaches 0 in layers 6-10, meaning that positional information is practically ignored. However, in layers 1-5, some of the attention heads keep a high value of 
$\sigma(\lambda_h)$, hence take advantage of positional information. Interestingly, the ConViT-Ti only uses positional information up to layer 4, whereas the ConViT-B uses it up to layer 6 (see App.~\ref{app:nonlocality}), suggesting that larger models - which are more under-specified - benefit more from the convolutional prior. These observations highlight the usefulness of the gating parameter in terms of interpretability.

\begin{figure}[h!]
    \centering
    \hspace{-1em}
    \begin{subfigure}[b]{.33\columnwidth}
    \includegraphics[width=\linewidth]{figs/photo_701.pdf}
    \caption{Input}
    \end{subfigure}   
    \hspace{-1.3em}
    \begin{subfigure}[b]{.68\columnwidth}
    \includegraphics[width=\linewidth]{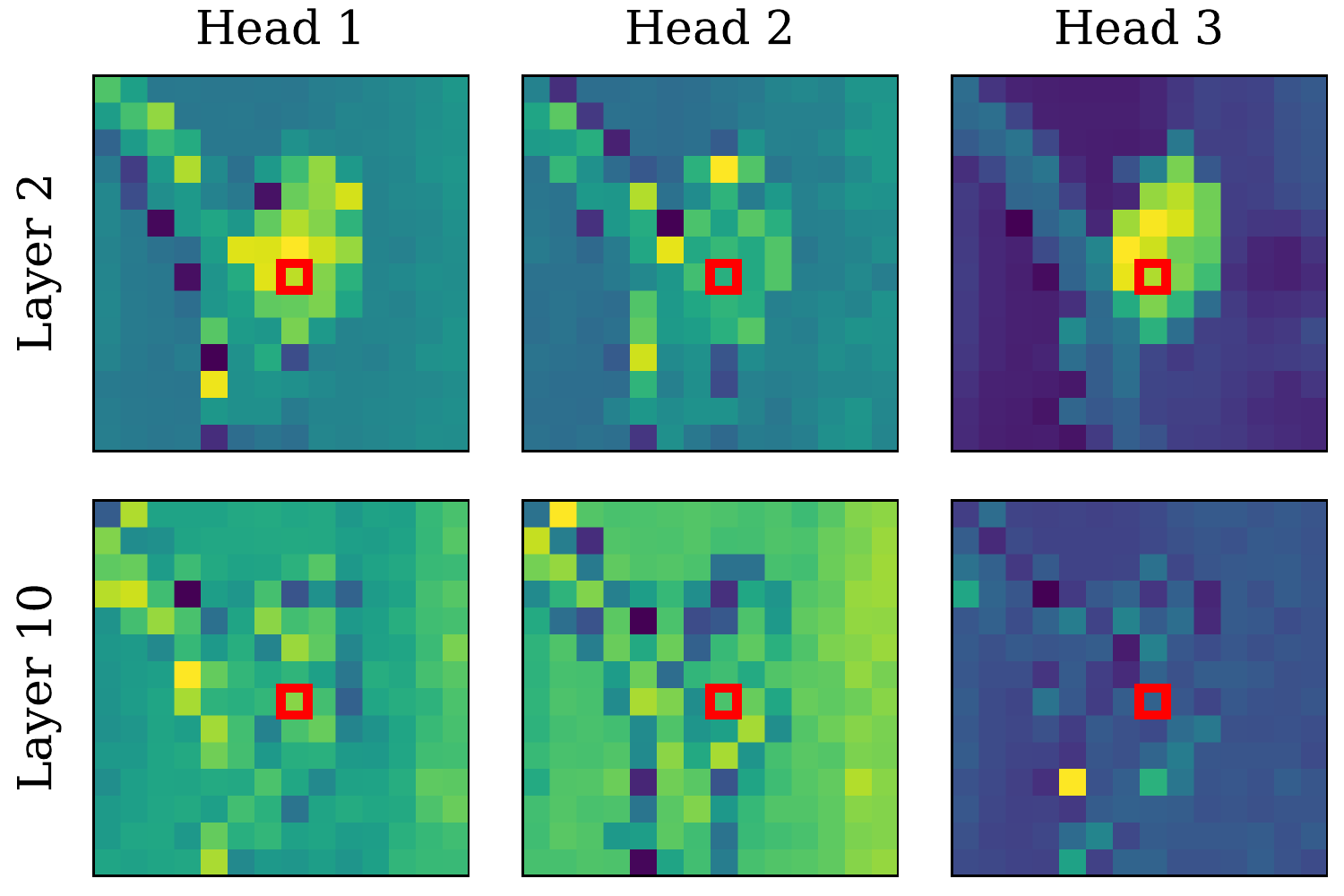}
    \caption{DeiT}
    \end{subfigure}
    \par\bigskip
    \begin{subfigure}[b]{.95\columnwidth}
    \includegraphics[width=\linewidth]{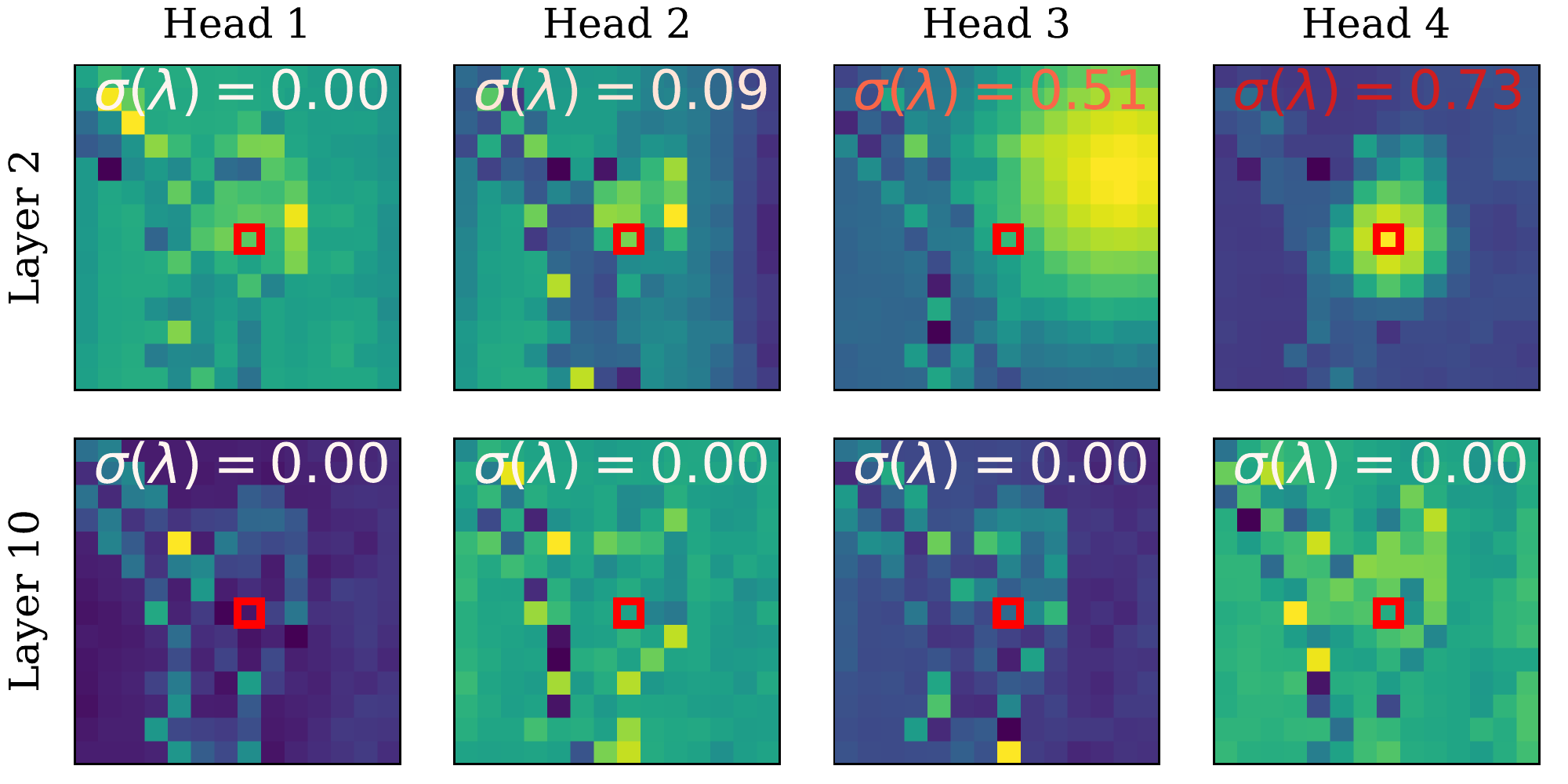}
    \caption{ConViT}
    \end{subfigure}
    \caption{\textbf{The ConViT learns more diverse attention maps}. \textit{Left:} input image which is embedded then fed into the models. The query patch is highlighted by a red box and the colormap is logarithmic to better reveal details. \textit{Center:} attention maps obtained by a DeiT-Ti after 300 epochs of training on ImageNet. \textit{Right:} Same for ConViT-Ti. In each map, we indicated the value of the gating parameter in a color varying from white (for heads paying attention to content) to red (for heads paying attention to position). Attention maps for more images and heads are shown in SM.~\ref{app:attention-maps}.}
    \label{fig:maps-final}
\end{figure}

The inner workings of the ConViT are further revealed by the attention maps of Fig.~\ref{fig:maps-final}, which are obtained by propagating an embedded input image through the layers and selecting a query patch at the center of the image\footnote{We do not show the attention paid to the class token in the SA layers}. In layer 10, (bottom row), the attention maps of DeiT and ConViT look qualitatively similar: they both perform content-based attention. In layer 2 however (top row), the attention maps of the ConViT are more varied: some heads pay attention to content (heads 1 and 2) whereas other focus mainly on position (heads 3 and 4). Among the heads which focus on position, some stay highly localized (head 4) whereas others broaden their attention span (head 3). The interested reader can find more attention maps in SM.~\ref{app:attention-maps}.

\begin{figure}[t]
    \centering
    \includegraphics[width=.36\columnwidth]{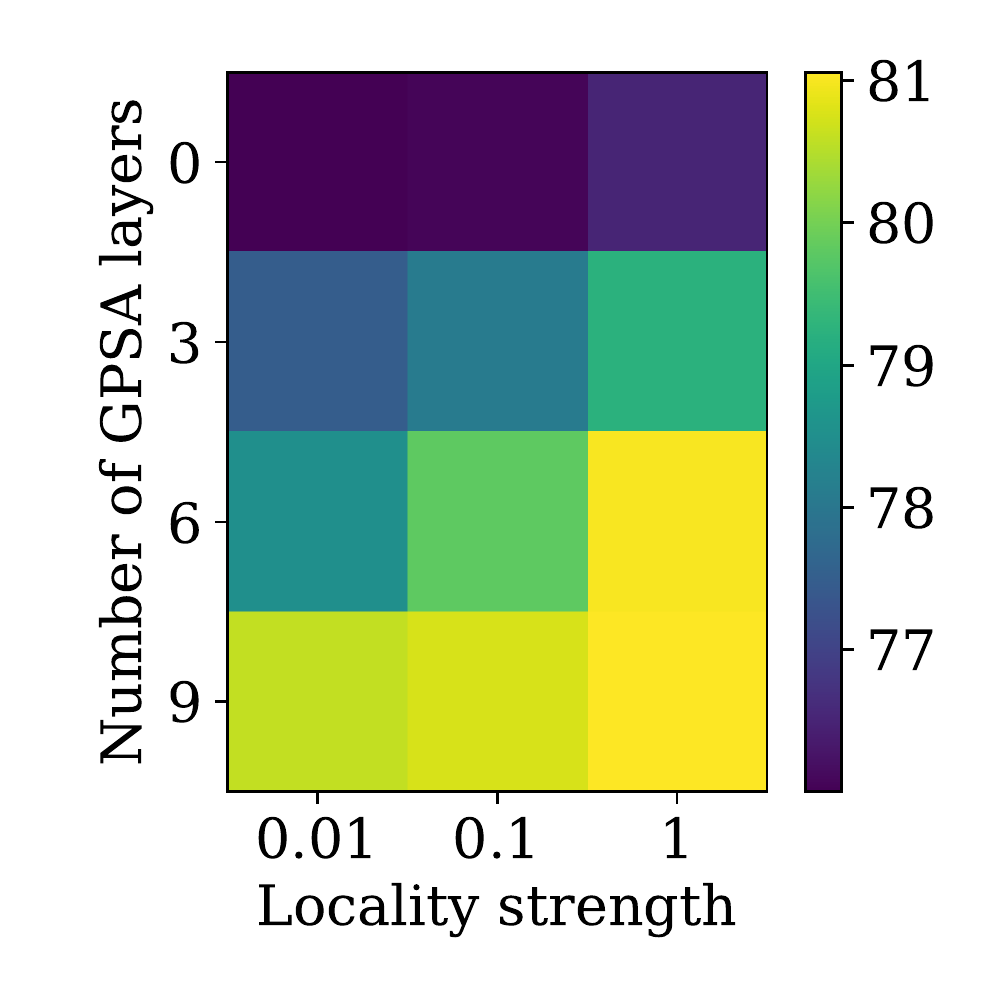}
    \includegraphics[width=.62\columnwidth]{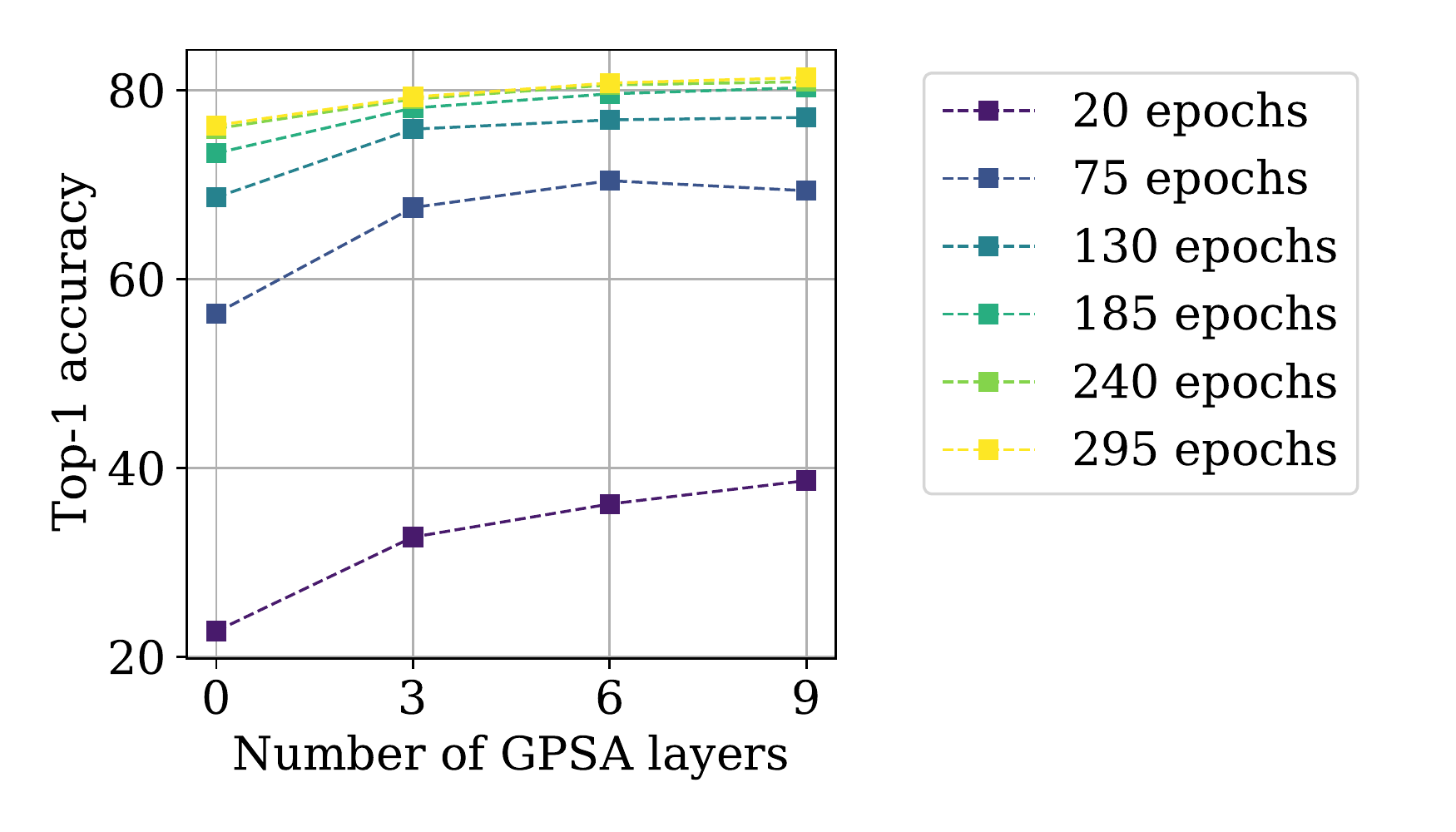}
    \caption{\textbf{The beneficial effect of locality}. \textit{Left:} As we increase the locality strength (i.e. how focused each attention head is its associated patch) and the number of GPSA layers of a ConViT-S+, the final top-1 accuracy increases significantly. \textit{Right:} The beneficial effect of locality is particularly strong in the early epochs. 
    }
    \label{fig:strength}
\end{figure}


\begin{table}[h]
    \centering
    \footnotesize
    \begin{tabular}{c|c|c|c|c|c|c}
    \toprule
        Ref. & \thead{ Train \\ gating} & \thead{Conv \\init} & \thead{Train \\GPSA} & \thead{Use\\GPSA} & \thead{Full\\ data} & \thead{10\%\\ data} \\
        \midrule
        a {\tiny (ConViT)} & \cmark & \cmark & \cmark & \cmark & \textbf{82.2} & \textbf{59.7}\\
        b & \xmark & \cmark &  \cmark & \cmark & 82.0 & 57.4\\
        c & \cmark & \xmark &  \cmark & \cmark & 81.4 & 56.9\\
        d & \xmark & \xmark &  \cmark & \cmark & 81.6 & 54.6\\
        e {\tiny (DeiT)} &  \xmark  & \xmark & \xmark & \xmark & 79.1 & 47.8\\
        f & \xmark & \cmark &  \xmark & \cmark & 78.6 & 54.3\\
        g & \xmark & \xmark &  \xmark & \cmark & 73.7 & 44.8\\
    \bottomrule
    \end{tabular}
    \caption{\textbf{Gating and convolutional initialization play nicely together.} We ran an ablation study on the ConViT-S+ trained for 300 epochs on the full ImageNet training set and on 10\% of the training data. From the left column to right column, we experimented freezing the gating parameters to 0, removing the convolutional initialization, freezing the GPSA layers and removing them altogether.}
    \label{tab:ablation}
\end{table}

\paragraph{Strong locality is desirable}

We next investigate how the performance of the ConViT is affected by two important hyperparameters of the ConViT: the \emph{locality strength}, $\alpha$, which determines how focused the heads are around their center of attention, and the number of SA layers replaced by GPSA layers. We examined the effects of these hyperparameters on ConViT-S, trained on the first 100 classes of ImageNet. As shown in Fig.~\ref{fig:strength}(a), final test accuracy increases both with the locality strength and with the number of GPSA layers; in other words, the more convolutional, the better.

In Fig.~\ref{fig:strength}(b), we show how performance at various stages of training is impacted by the presence of GPSA layers. We see that the boost due to GPSA is particularly strong during the early stages of training: after 20 epochs, using 9 GPSA layers leads the test-accuracy to almost double, suggesting that the convolution initialization gives the model a substantial ``head start". This speedup is of practical interest in itself, on top of the boost in final performance.

\paragraph{Ablation study}
In Tab.~\ref{tab:ablation}, we present an ablation on the ConViT, denoted as \textbf{[a]}. We experiment removing the positional gating \textbf{[b]}\footnote{To remove gating, we freeze all gating parameters to $\lambda=0$ so that the same amount of attention is paid to content and position.}, the convolutional initialization \textbf{[c]}, both gating and the convolutional initialization \textbf{[d]}, and the GPSA altogether (\textbf{[e]}, which leaves us with a plain DeiT). 

Surprisingly, on full ImageNet, GPSA without gating \textbf{[d]} already brings a substantial benefit over the DeiT (+2.5), which is mildly increased by the convolutional initialization (\textbf{[b]}, +2.9). As for gating, it helps a little in presence of the convolutional initialization (\textbf{[a]}, +3.1), and is unhelpful otherwise. These mild improvements due to gating and convolutional initialization (likely due to performance saturation above 80\% top-1) become much clearer in the low data regime. Here, GPSA alone brings +6.8, with an extra +2.3 coming from gating, +2.8 from convolution initialization and +5.1 with the two together, illustrating their complementarity. 

We also investigated the performance of the ConViT with all GPSA layers frozen, leaving only the FFNs to be trained in the first 10 layers. As one could expect, performance is strongly degraded in the full data regime if we initialize the GPSA layers randomly (\textbf{[f]}, -5.4 compared to the DeiT). However, the convolutional initialization remarkably enables the frozen ConViT to reach a very decent performance, almost equalling that of the DeiT (\textbf{[e]}, -0.5). In other words, replacing SA layers by random ``convolutions" hardly impacts performance. In the low data regime, the frozen ConViT even outperforms the DeiT by a margin (+6.5). This naturally begs the question: is attention really key to the success of ViTs~\cite{dong2021attention,tolstikhin2021mlp,touvron2021resmlp}?

\section{Conclusion and perspectives}

The present work investigates the importance of initialization and inductive biases in learning with vision transformers. By showing that one can take advantage of convolutional constraints in a soft way, we merge the benefits of architectural priors and expressive power. The result is a simple recipe that improves trainability and sample efficiency, without increasing model size or requiring any tuning. 

Our approach can be summarized as follows: instead of interleaving convolutional layers with SA layers as done in hybrid models, let the layers decide whether to be convolutional or not by adjusting a set of gating parameters. More generally, combining the biases of varied architectures and letting the model choose which ones are best for a given task could become a promising direction, reducing the need for greedy architectural search while offering higher interpretability. 

Another direction which will be explored in future work is the following: if SA layers benefit from being initialized as random convolutions, could one reduce even more drastically their sample complexity by initializing them as pre-trained convolutions?




\paragraph{Acknowledgements}
We thank Hervé Jégou and Francisco Massa for helpful discussions. SD and GB acknowledge funding from the French government under management of Agence Nationale de la Recherche as part of the “Investissements d’avenir” program, reference ANR-19-P3IA-0001 (PRAIRIE 3IA Institute). 






\bibliography{refs}
\bibliographystyle{icml2020}

\clearpage
\appendix
\clearpage
\onecolumn

\section{The importance of positional gating}
\label{app:without-gpsa}

\begin{figure}[htb]
    \centering
    \includegraphics[width=.4\columnwidth]{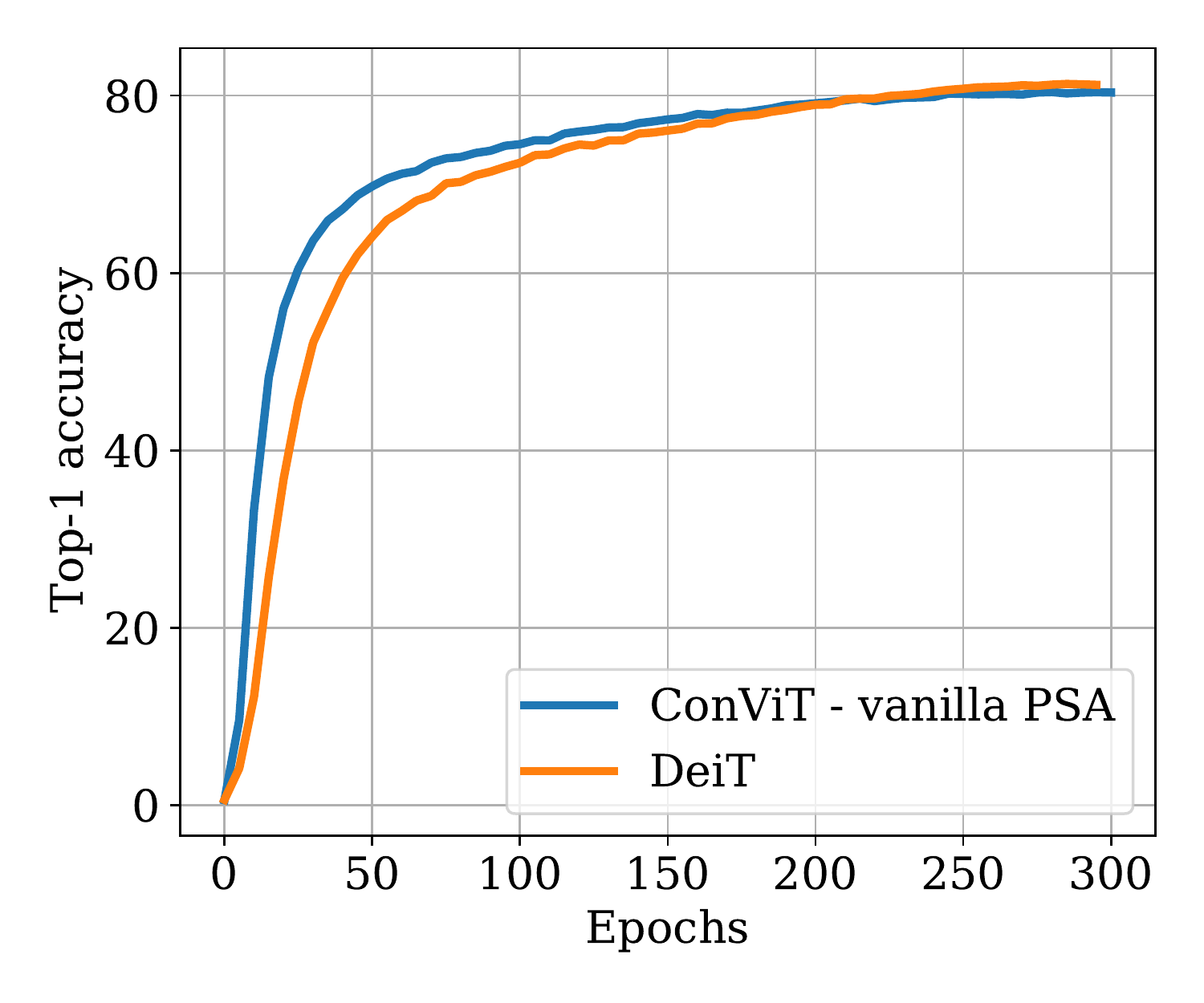}
    \caption{\textbf{Convolutional initialization without GPSA is helfpul during early training but deteriorates final performance.} We trained the ConViT-B along with its DeiT-B counterpart for 300 epochs on ImageNet, replacing the GPSA layers of the ConViT-B by vanilla PSA layers.}
    \label{fig:vanilla-psa}
\end{figure}

In the main text, we discussed the importance of using GPSA layers instead of the standard PSA layers, where content and positional information are summed before the softmax and lead the attention heads to focus only on the positional information. 
We give evidence for this claim in Fig.~\ref{fig:vanilla-psa}, where we train a ConViT-B for 300 epochs on ImageNet, but replace the GPSA by standard PSA. The convolutional initialization of the PSA still gives the ConViT a large advantage over the DeiT baseline early in training. However, the ConViT stays in the convolutional configuration and ignores the content information, as can be seen by looking at the attention maps (not shown). Later in training, the DeiT catches up and surpasses the performance of the ConViT by utilizing content information.

\section{The effect of distillation}
\label{app:distillation}

\paragraph{Nonlocality}

\begin{figure}[htb]
    \centering
    \includegraphics[width=.8\linewidth]{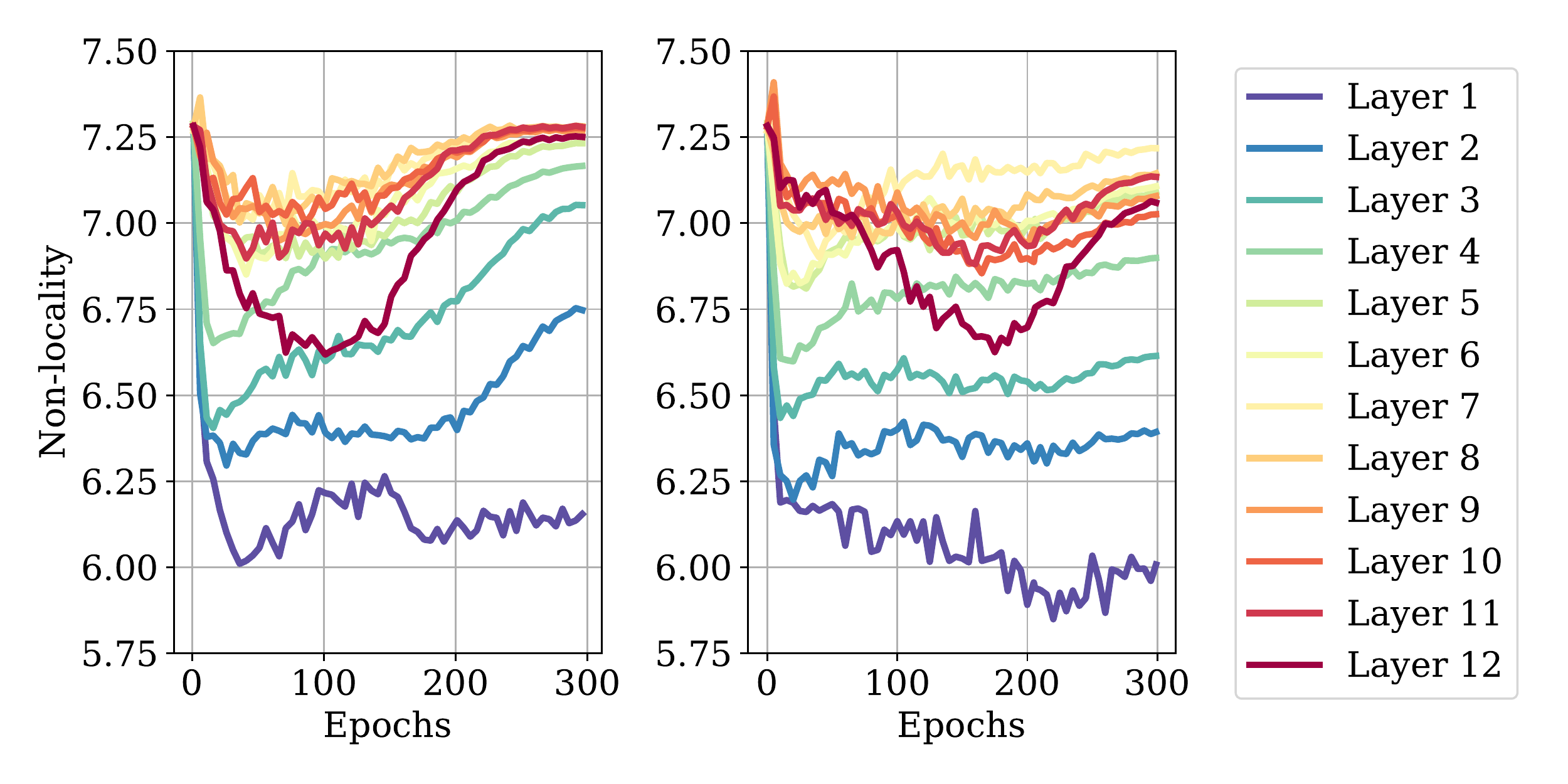}
    \caption{\textbf{Distillation pulls the DeiT towards a more local configuration.} We plotted the nonlocality metric defined in Eq.~\ref{eq:nonlocality} throughout training, for the DeiT-S trained on ImageNet. \textit{Left:} regular training. \textit{Right:} training with hard distillation from a RegNet teacher, by means of the distillation introduced in~\citep{touvron2020training}.}
    \label{fig:nonlocality-distillation}
\end{figure}

In Fig.~\ref{fig:nonlocality-distillation}, we compare the nonlocality curves of Fig.~\ref{fig:nonlocality} of the main text with those obtained when the DeiT is trained via hard distillation from a RegNetY-16GF (84M parameters)~\cite{radosavovic2020designing}, as in~\citet{touvron2020training}. In the distillation setup, the nonlocality still drops in the early epochs of training, but increases less at late times compared to without distillation. Hence, the final internal states of the DeiT are more ``local" due to the distillation. This suggests that knowledge distillation transfers the locality of the convolutional teacher to the student, in line with the results of~\cite{abnar2020transferring}.

\paragraph{Performance}

The hard distillation introduced in~\citet{touvron2020training} greatly improves the performance of the DeiT. We have verified the complementarity of their distillation methods with our ConViT. In the same way as in the DeiT paper, we used a RegNet-16GF teacher and experimented hard distillation during 300 epochs on ImageNet. The results we obtain are summarized in Tab.~\ref{tab:distillation}.

\begin{table}[h]
    \centering
    \begin{tabular}{c|cccc}
    \toprule
    Method & DeiT-S (22M) & DeiT-B (86M) & ConViT-S+ (48M)\\
    \midrule
    No distillation &  79.8 & 81.8 &\textbf{82.2}\\
    Hard distillation  & 80.9 & \textbf{83.0} & 82.9\\ 
    \bottomrule
    \end{tabular}
    \caption{Top-1 accuracies of the ConViT-S+ compared to the DeiT-S and DeiT-B, both trained for 300 epochs on ImageNet.}
    \label{tab:distillation}
\end{table}
Just like the DeiT, the ConViT benefits from distillation, albeit somewhat less than the DeiT, as can be seen from the DeiT-B performing less well than the ConViT-S+ without distillation but better with distillation. This hints to the fact that the convolutional inductive bias transferred from the teacher is redundant with its own convolutional prior.

Nevertheless, the performance improvement obtained by the ConViT with hard distillation demonstrates that instantiating soft inductive biases directly in a model can yield benefits on top of those obtained by instantiating such biases indirectly, in this case via distillation.

\section{Further performance results}
\label{app:performance}

\begin{figure*}[h]
    \centering
    \begin{subfigure}[b]{.31\textwidth}
    \includegraphics[width=\linewidth]{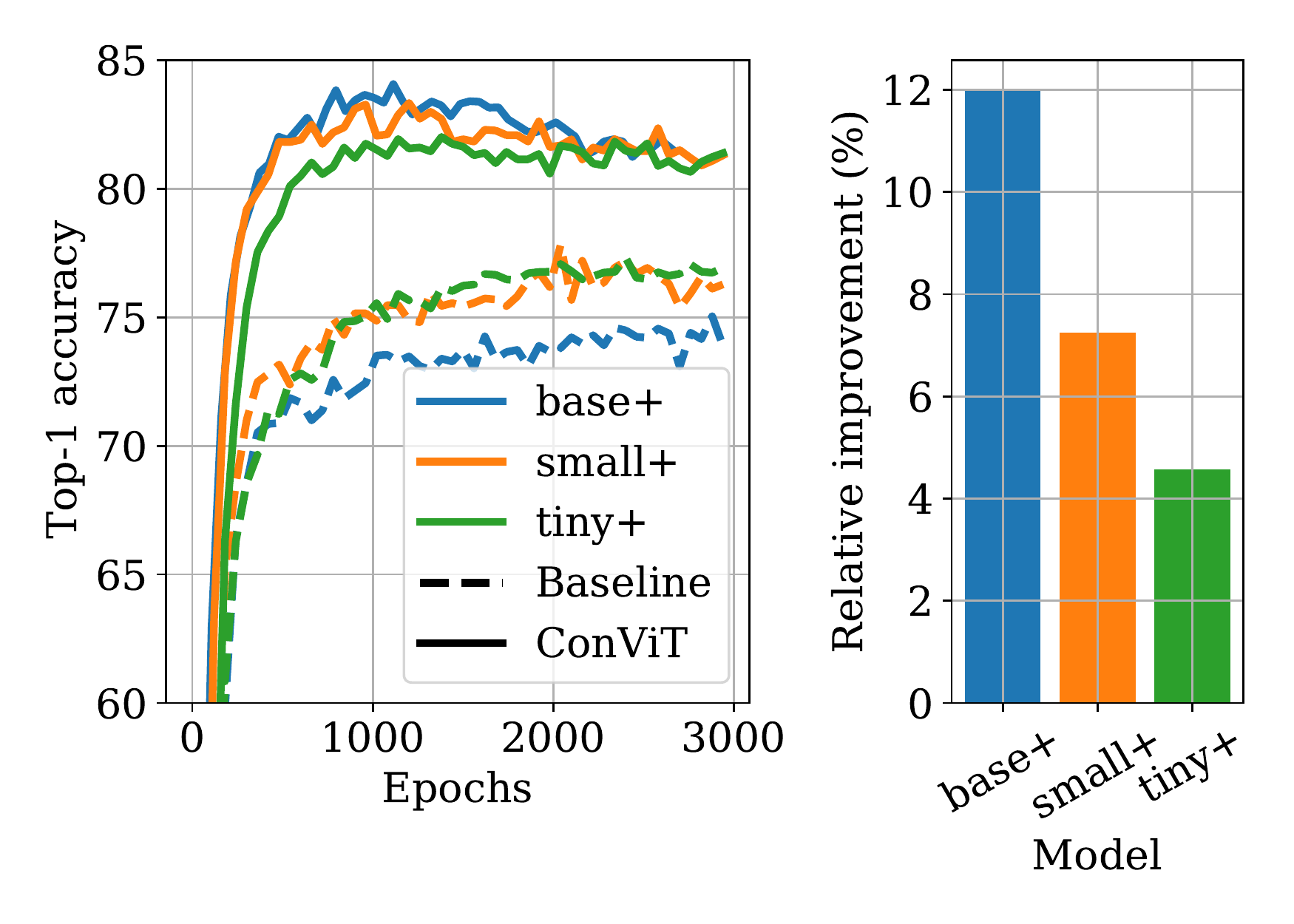}
    \caption{CIFAR100}
    \end{subfigure}
    \begin{subfigure}[b]{.31\textwidth}
    \includegraphics[width=\linewidth]{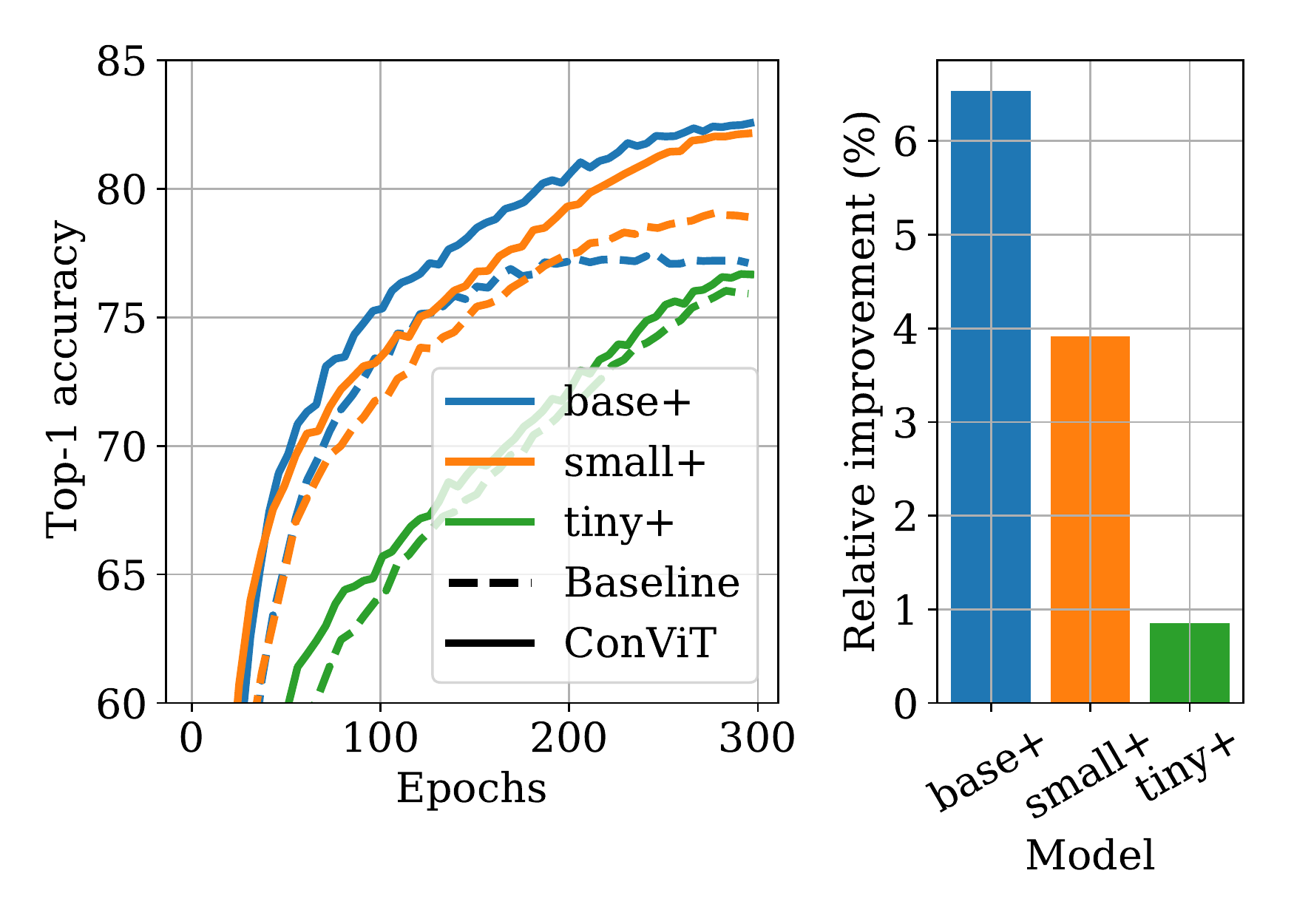}
    \caption{ImageNet-1k}
    \end{subfigure}
    \begin{subfigure}[b]{.35\textwidth}
    \includegraphics[width=\linewidth]{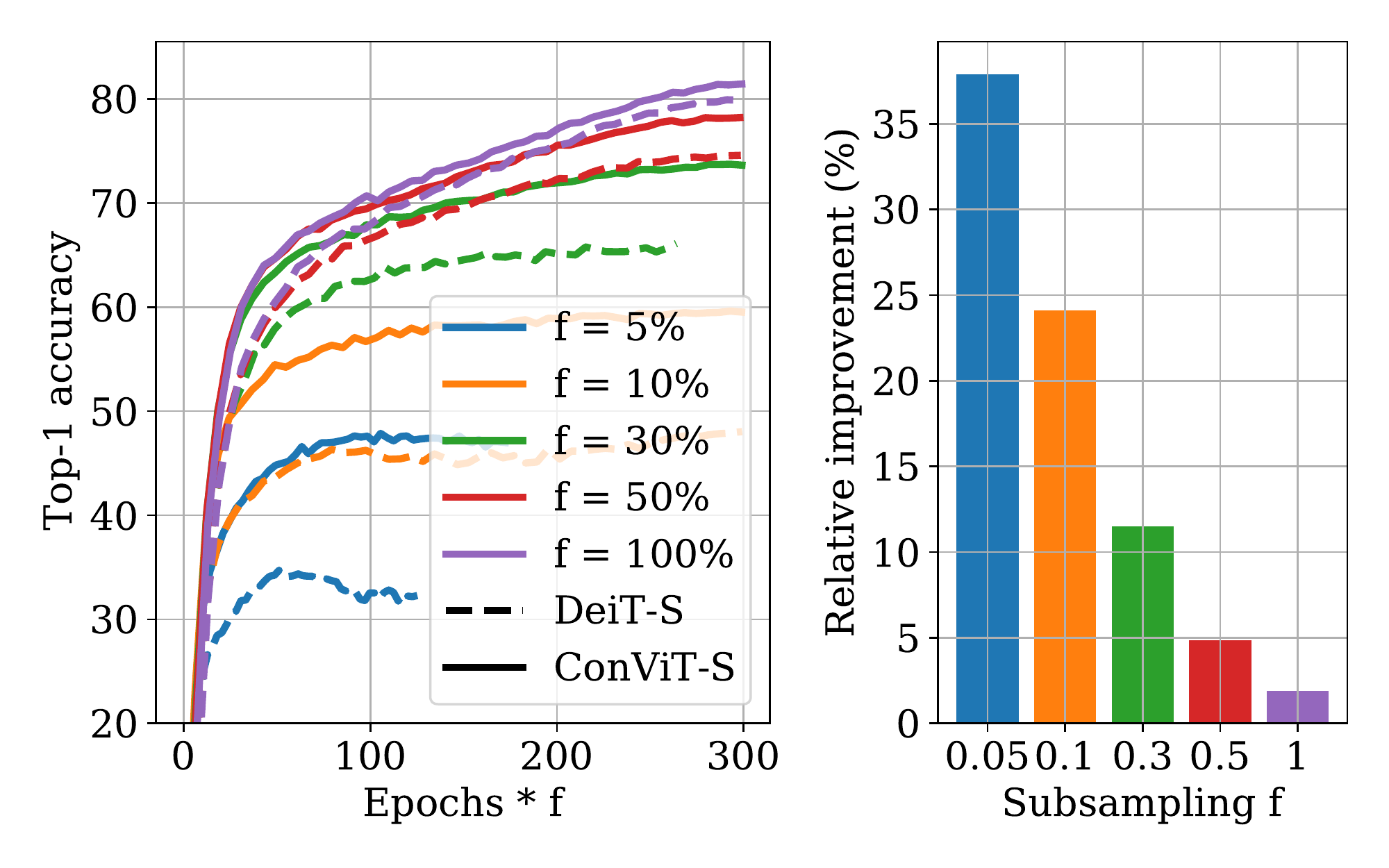}
    \caption{Subsampled ImageNet}
    \end{subfigure}
    \caption{\textbf{The convolutional inductive bias is particularly useful for large models applied to small datasets.} Each of the three panels displays the top-1 accuracy of the ConViT+ model and their corresponding DeiT+ throughout training, as well as the relative improvement between the best top-1 accuracy reached by the DeiT+ and that reached by the ConViT+. \textit{Left:} tiny, small and base models trained for 3000 epochs on CIFAR100. \textit{Middle:} tiny, small and base models trained for 300 epochs on ImageNet-1k. The relative improvement of the ConViT over the DeiT increases with model size. \textit{Right:} small model trained on a subsampled version of ImageNet-1k, where we only keep a fraction $f\in\{0.05,0.1,0.3,0.5,1\}$ of the images of each class. The relative improvement of the ConViT over the DeiT increases as the dataset becomes smaller.}
    \label{fig:time-dependence}
\end{figure*}

In Fig.~\ref{fig:time-dependence}, we display the time evolution of the top-1 accuracy of our ConViT+ models on CIFAR100, ImageNet and subsampled ImageNet, along with a comparison with the corresponding DeiT+ models. 

For CIFAR100, we kept all hyperparameters unchanged, but rescaled the images to $224\times 224$ and increased the number of epochs (adapting the learning rate schedule correspondingly) to mimic the ImageNet scenario. After 1000 epochs, the ConViTs shows clear signs of overfitting, but reach impressive performances (82.1\% top-1 accuracy with 10M parameters, which is better than the EfficientNets reported in ~\cite{zhao2020splitnet}).

\begin{figure}[t]
    \centering
    \includegraphics[width=.35\columnwidth]{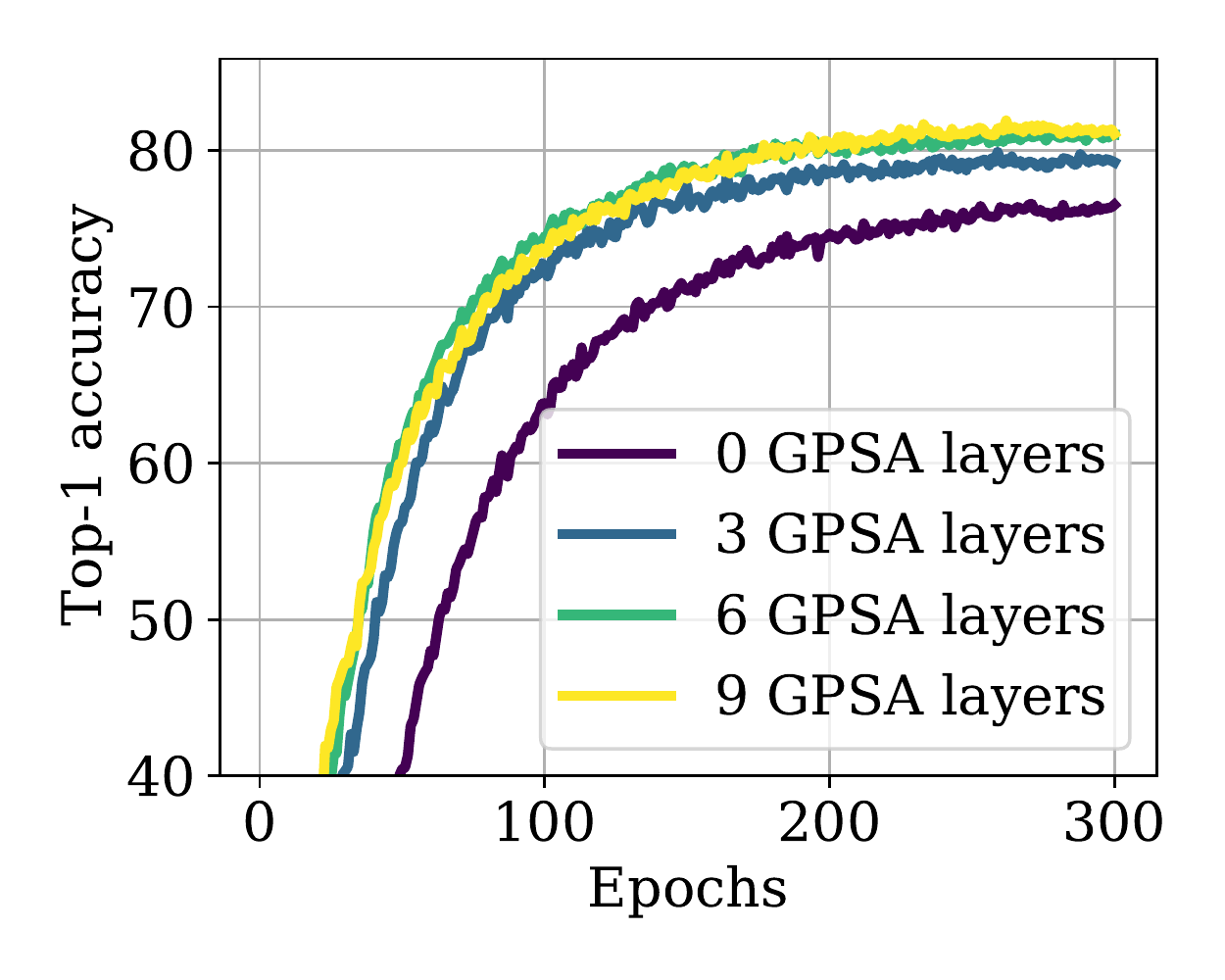}
    \includegraphics[width=.62\columnwidth]{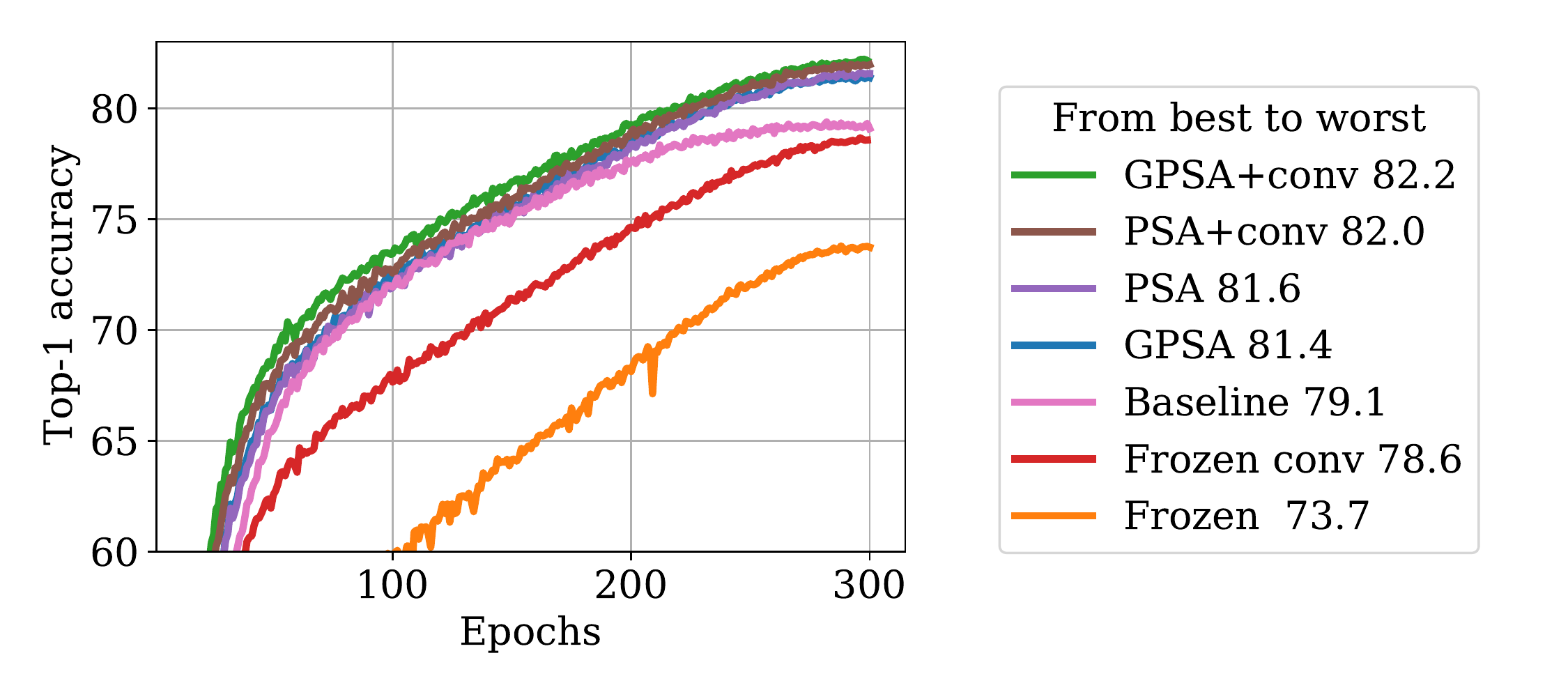}
    \caption{\textbf{Impact of various ingredients of the ConViT on the dynamics of learning.} In both cases, we train the ConViT-S+ for 300 epochs on first 100 classes of ImageNet. \textit{Left:} ablation on number of GPSA layers, as in Fig.~\ref{fig:strength}. \textit{Right:} ablation on various ingredients of the ConViT, as in Tab.~\ref{tab:ablation}. The baseline is the DeiT-S+ (pink). We experimented (i) replacing the 10 first SA layers by GPSA layers (``GPSA'') (ii) freezing the gating parameter of the GPSA layers (``frozen gate''); (iii) removing the convolutional initialization (``conv''); (iv) freezing all attention modules in the GPSA layers (``frozen''). The final top-1 accuracy of the various models trained is reported in the legend.}
    \label{fig:ablation-dynamics}
\end{figure}

In Fig.~\ref{fig:ablation-dynamics}, we study the impact of the various ingredients of the ConViT (presence and number of GPSA layers, gating parameters, convolutional initialization) on the dynamics of learning. 

\section{Effect of model size}
\label{app:nonlocality}

\begin{figure}[h]
    \begin{subfigure}[b]{\linewidth}
    \centering
    \includegraphics[width=.7\linewidth]{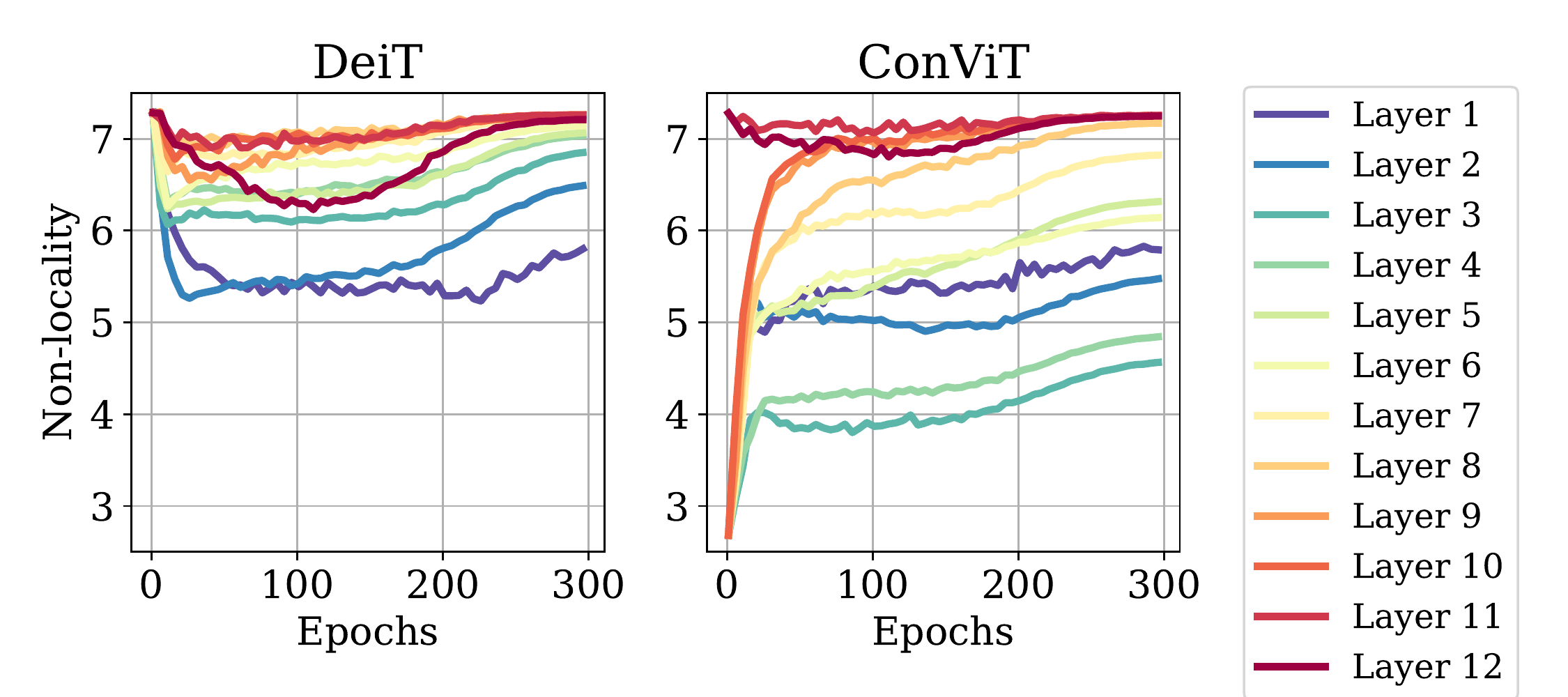}
    \caption{DeiT-Ti and ConViT-Ti}
    \end{subfigure}
    \begin{subfigure}[b]{\linewidth}
    \centering
    \includegraphics[width=.7\linewidth]{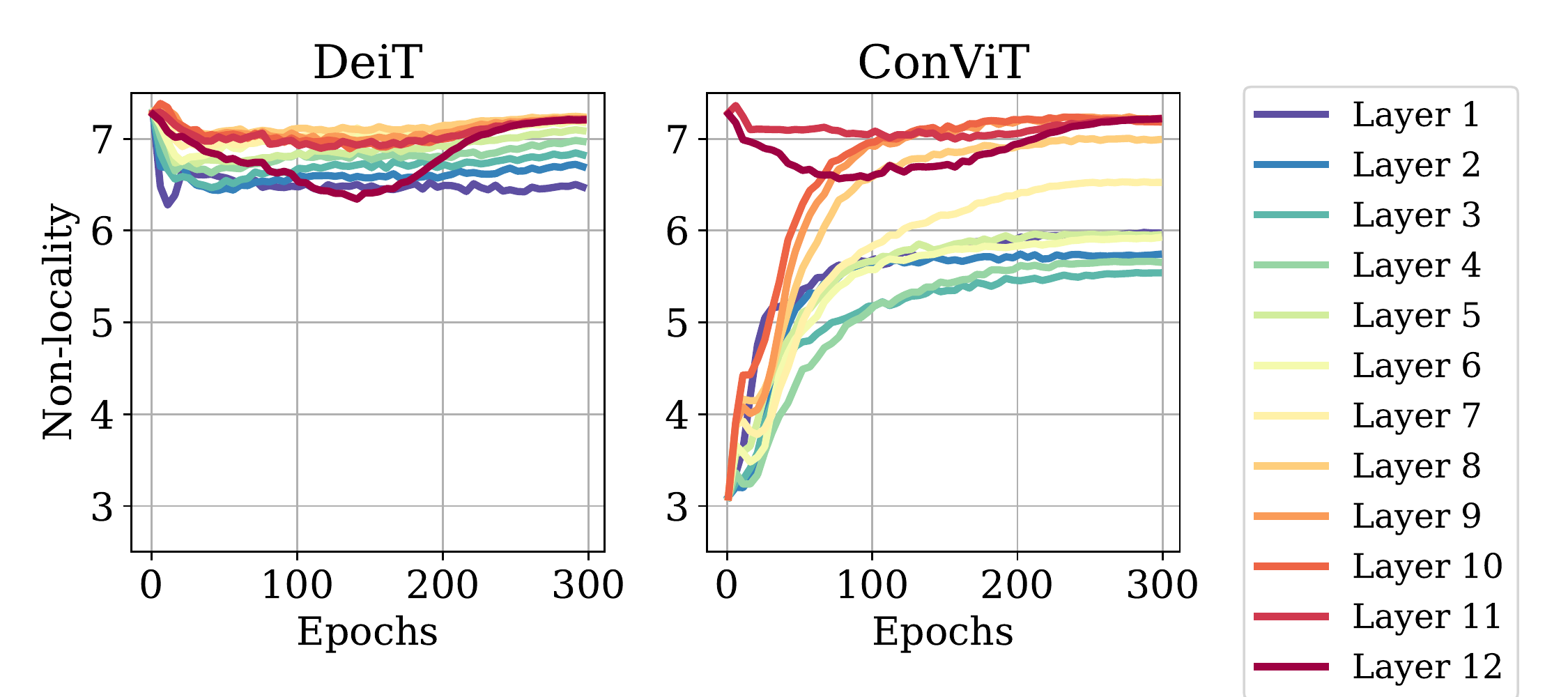}
    \caption{DeiT-B and ConViT-B}
    \end{subfigure}
    \caption{\textbf{The bigger the model, the more non-local the attention.} We plotted the nonlocality metric defined in Eq.~\ref{eq:nonlocality} of the main text (the higher, the further the attention heads look from the query pixel) throughout 300 epochs of training on ImageNet-1k. }
    \label{fig:nonlocality-sizes}
\end{figure}

In Fig.~\ref{fig:nonlocality-sizes}, we show the analog of Fig.~\ref{fig:nonlocality} of the main text for the tiny and base models. Results are qualitatively similar to those observed for the small model. Interestingly, the first layers of DeiT-B and ConViT-B reach significantly higher nonlocality than those of the DeiT-Ti and ConViT-Ti.

\begin{figure}[h]
    \begin{subfigure}[b]{\linewidth}
    \centering
    \includegraphics[width=.7\linewidth]{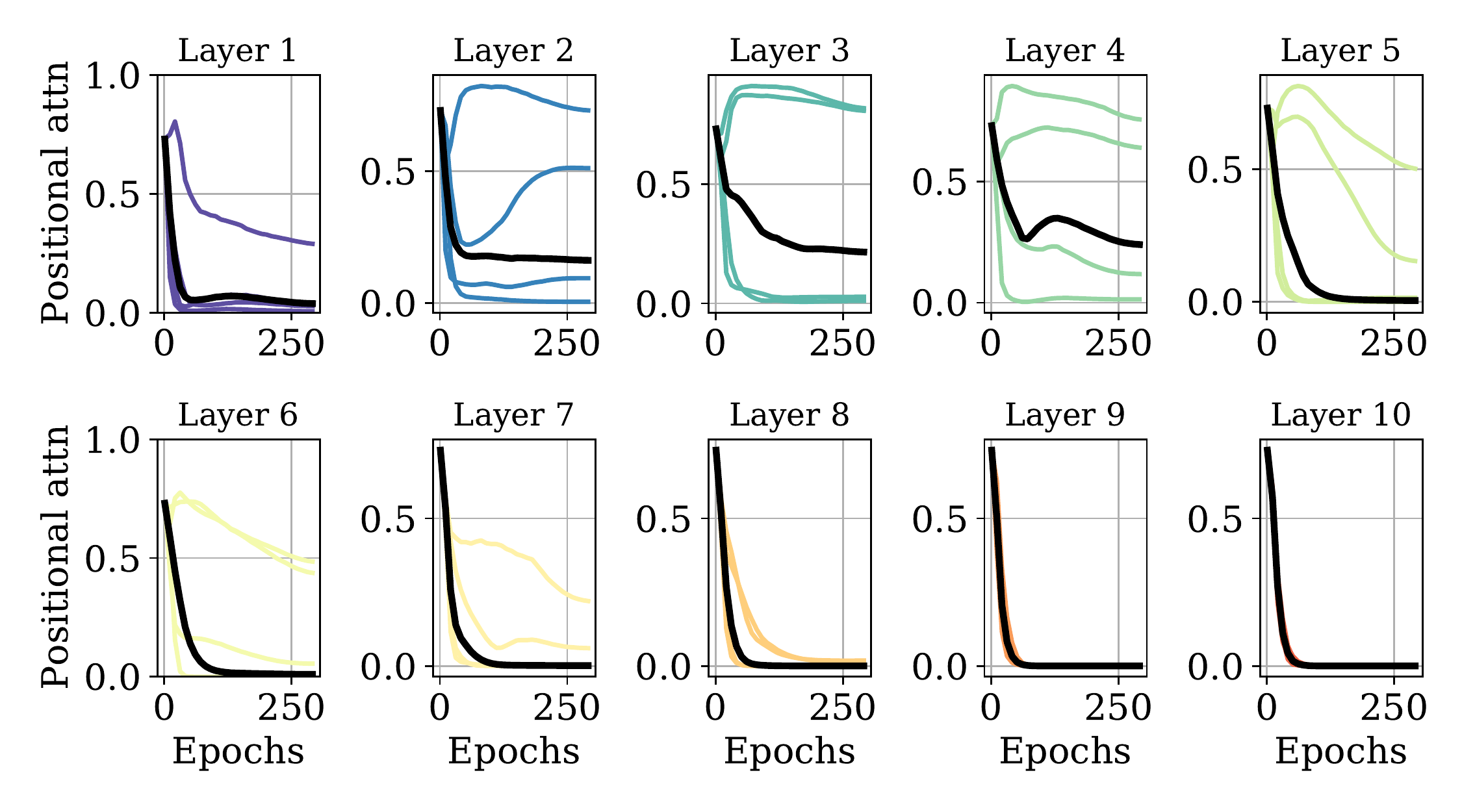}
    \caption{ConViT-Ti}
    \end{subfigure}
    \begin{subfigure}[b]{\linewidth}
    \centering
    \includegraphics[width=.7\linewidth]{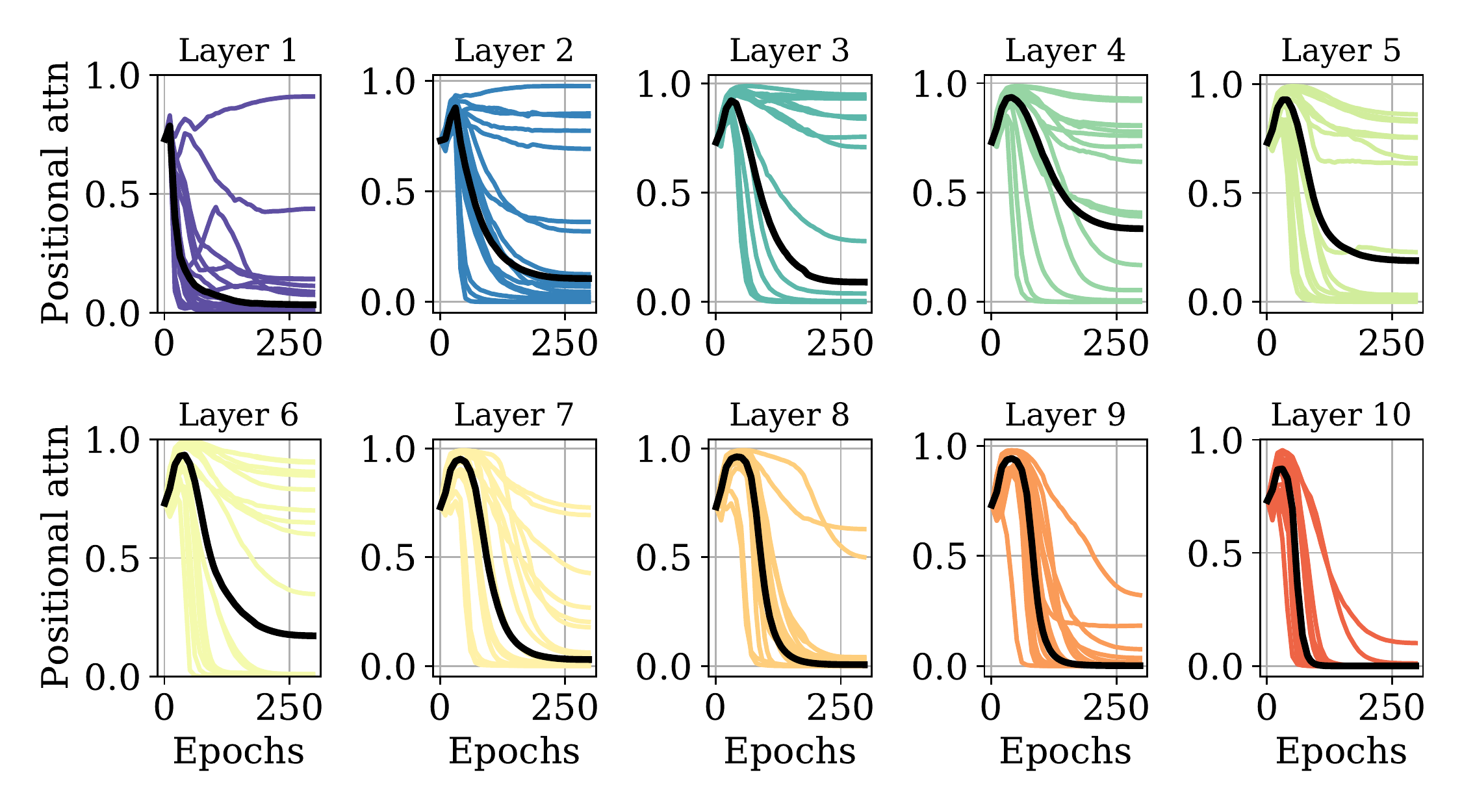}
    \caption{ConViT-B}
    \end{subfigure}
    \caption{\textbf{The bigger the model, the more layers pay attention to position.} We plotted the gating parameters of various heads and various layers, as in Fig.~\ref{fig:gating} of the main text (the lower, the less attention is paid to positional information) throughout 300 epochs of training on ImageNet-1k. Note that the ConViT-Ti only has 4 attention heads whereas the ConViT-B has 16, hence the different number of curves.}
    \label{fig:gating-sizes}
\end{figure}

In Fig.~\ref{fig:gating-sizes}, we show the analog of Fig.~\ref{fig:gating} of the main text for the tiny and base models. Again, results are qualitatively similar: the average weight of the positional attention, $\E_h\sigma(\lambda_h)$, decreases over time, so that more attention goes to the content of the image. Note that in the ConViT-Ti, only the first 4 layers still pay attention to position at the end of training (average gating parameter smaller than one), whereas for ConViT-S, the 5 first layers still do, and for the ConViT-B, the 6 first layers still do. This suggests that the larger (i.e. the more underspecified) the model is, the more layers make use of the convolutional prior.

\clearpage

\section{Attention maps}
\label{app:attention-maps}

\paragraph{Attention maps of the DeiT reveal locality}

\begin{figure}[htb]
    \centering
    \begin{subfigure}[b]{.49\columnwidth}
    \includegraphics[width=\linewidth]{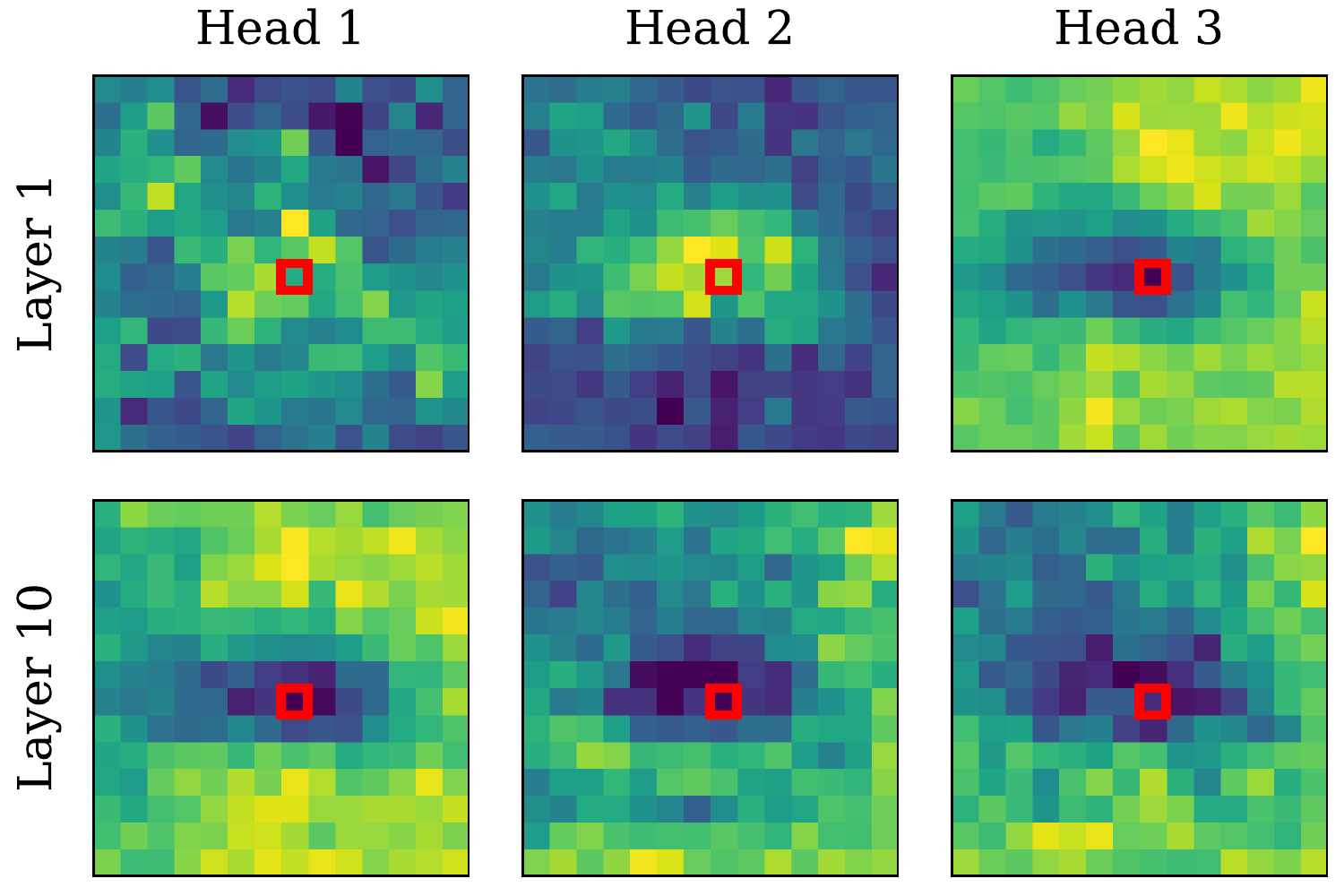}
    \caption{Before training}
    \end{subfigure}
    \begin{subfigure}[b]{.49\columnwidth}
    \includegraphics[width=\linewidth]{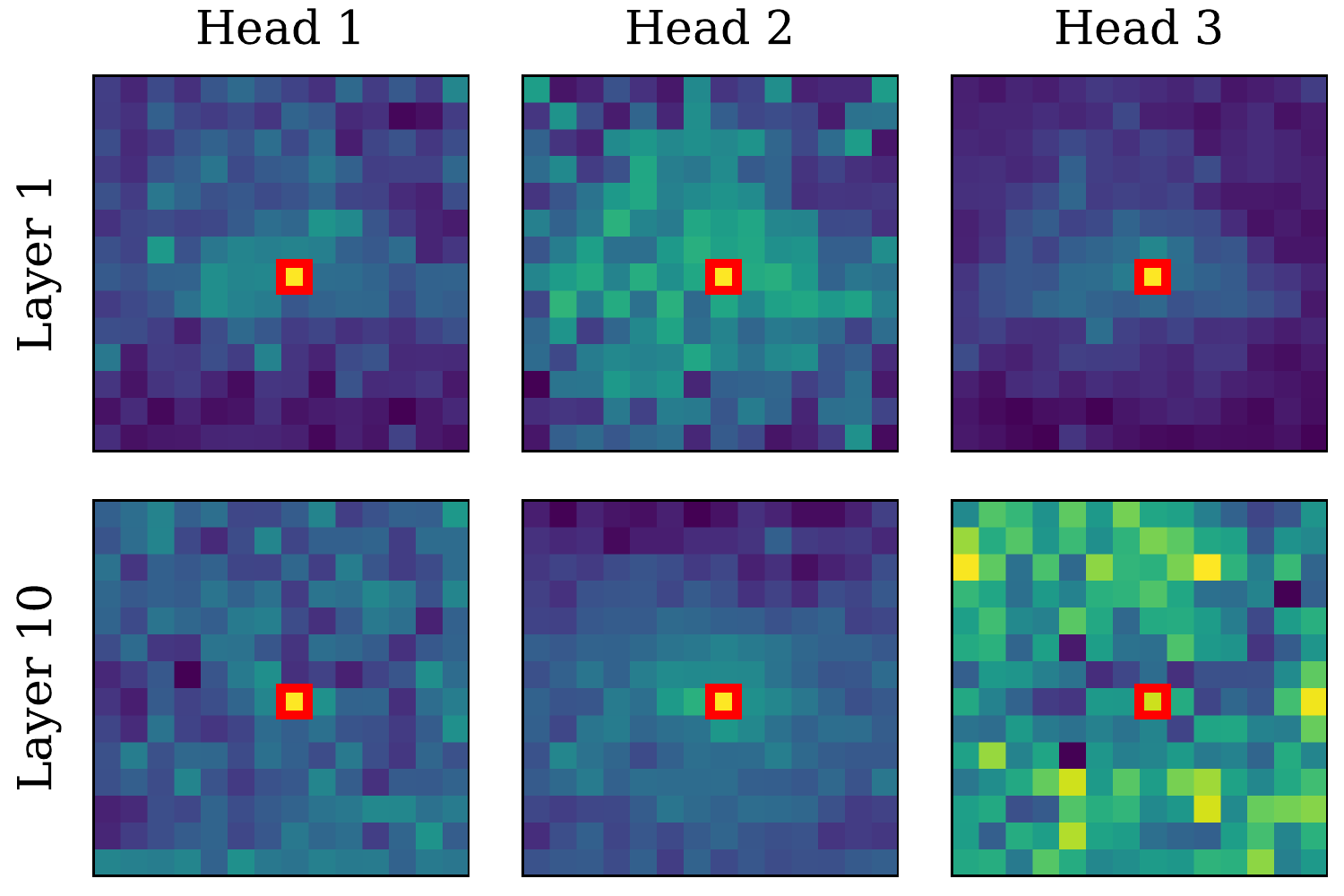}
    \caption{After training}
    \end{subfigure}
    \caption{\textbf{The averaged attention maps of the DeiT reveal locality at the end of training.} To better visualise the center of attention, we averaged the attention maps over 100 images. \textit{Top:} before training, the attention patterns exhibit a random structure. \textit{Bottom:} after training, most of the attention is devoted to the query pixel, and the rest is focused on its immediate surroundings.}
    \label{fig:attn-nonlocal}
\end{figure}

In Fig.~\ref{fig:attn-nonlocal}, we give some visual evidence for the fact that vanilla SA layers extract local information by averaging the attention map of the first and tenth layer of the DeiT over 100 images. Before training, the maps look essentially random. After training, however, most of the attention heads of the first layer focus on the query pixel and its immediate surroundings, whereas the attention heads of the tenth layer capture long-range dependencies.

\paragraph{Attention maps of the ConViT reveal the diversity of the attention heads}

\begin{figure*}[htb]
    \centering
    \includegraphics[width=.25\linewidth]{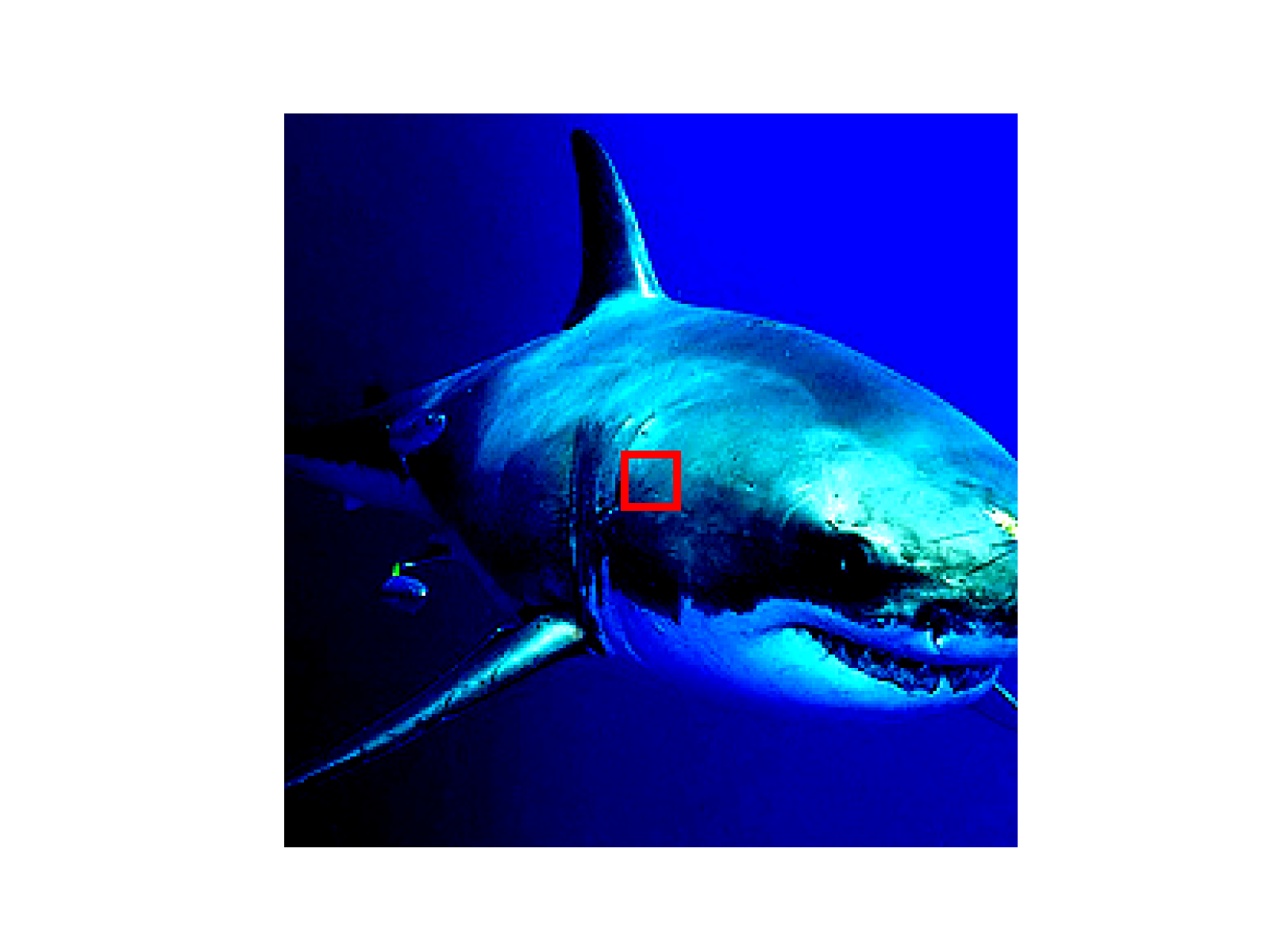}
    \includegraphics[width=.3\linewidth]{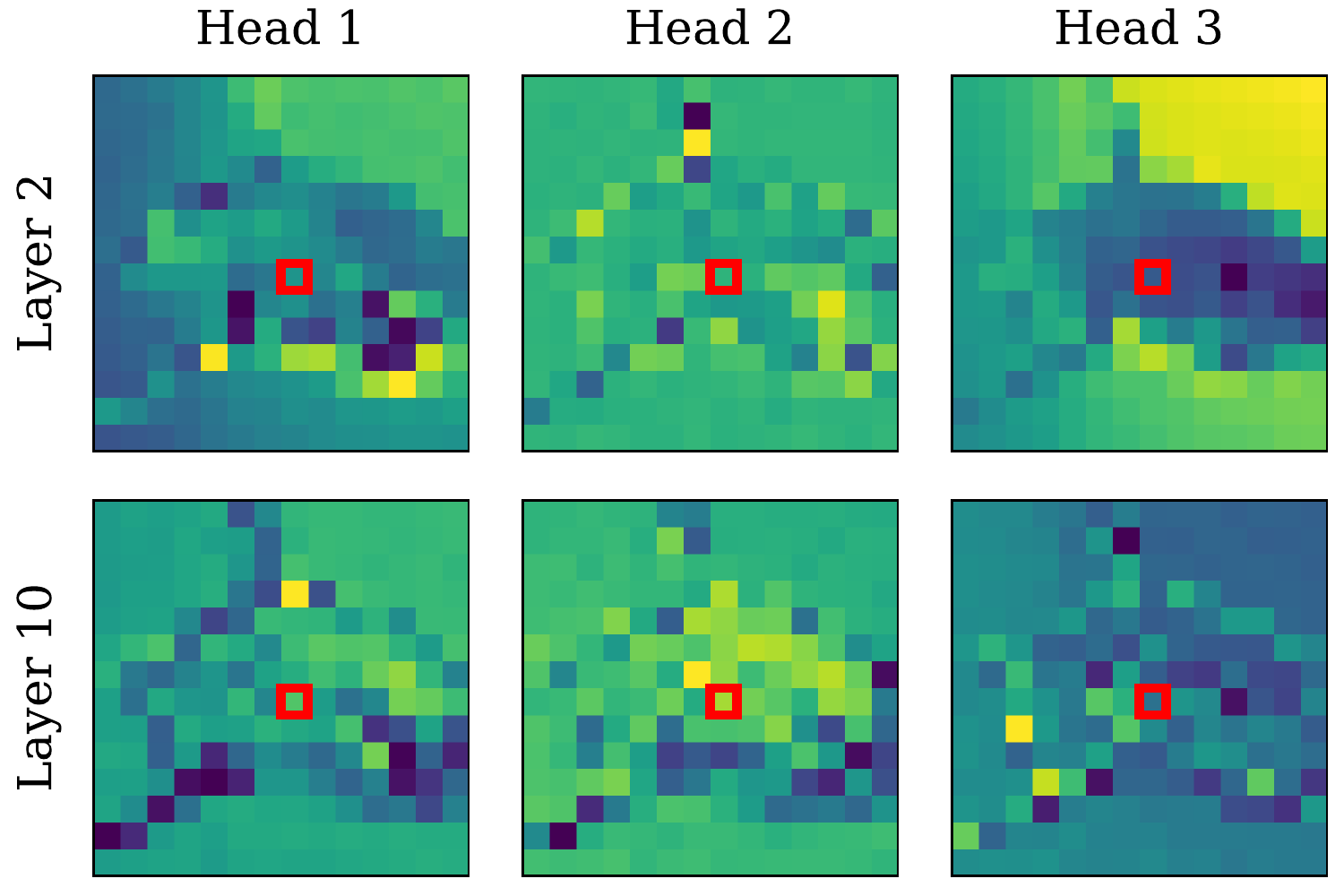}
    \hfill
    \includegraphics[width=.4\linewidth]{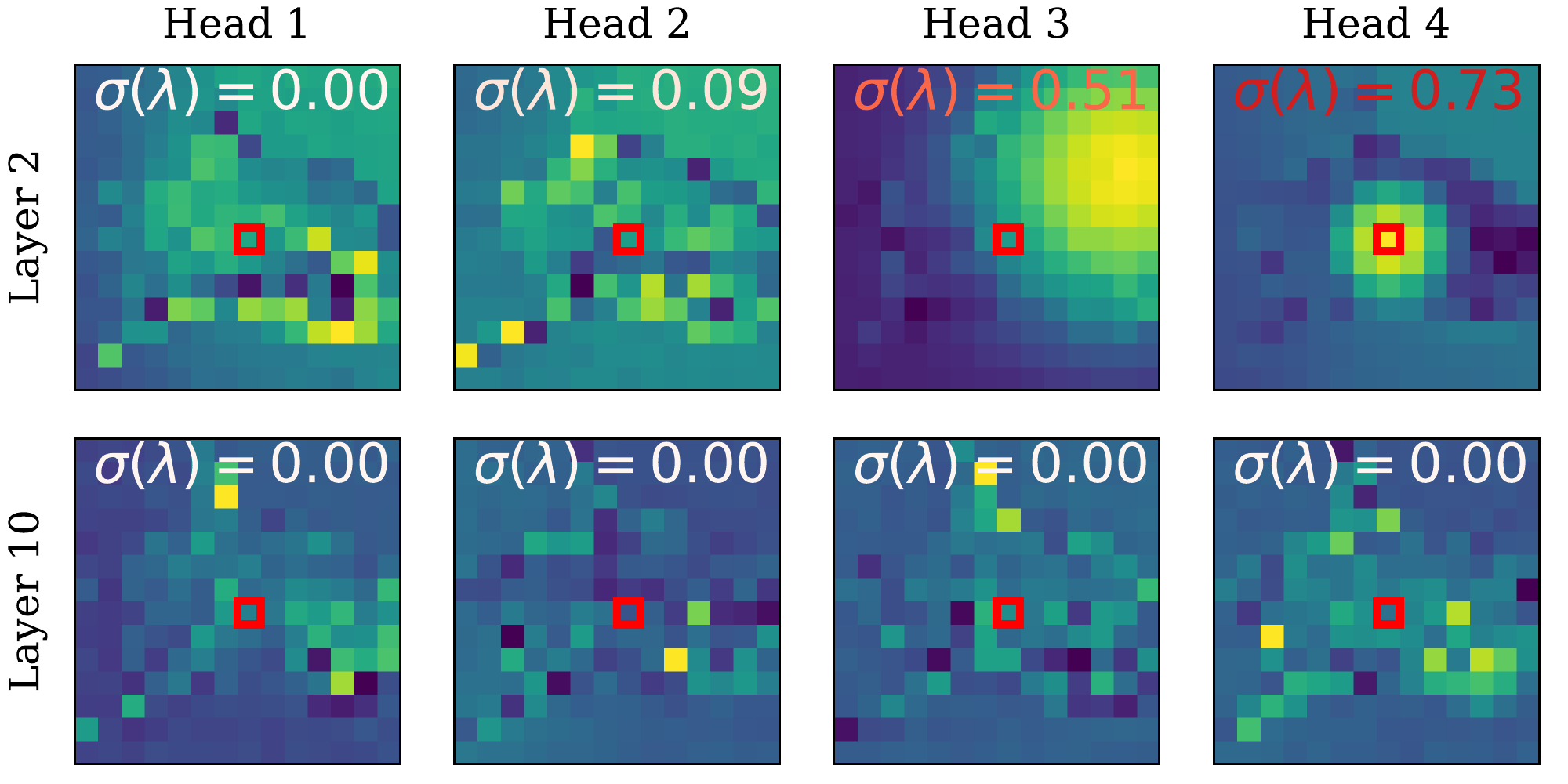}
    \includegraphics[width=.25\linewidth]{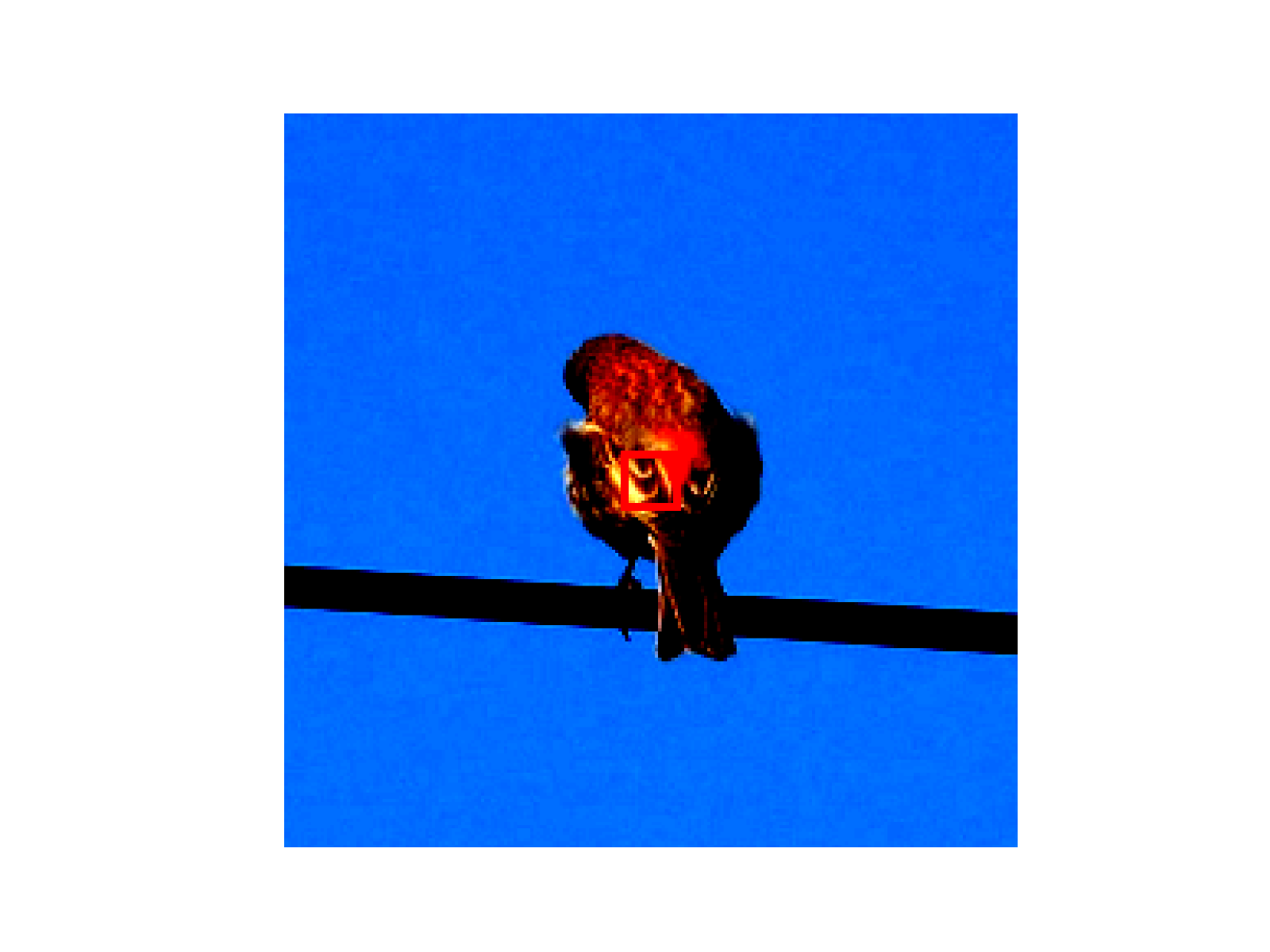}
    \includegraphics[width=.3\linewidth]{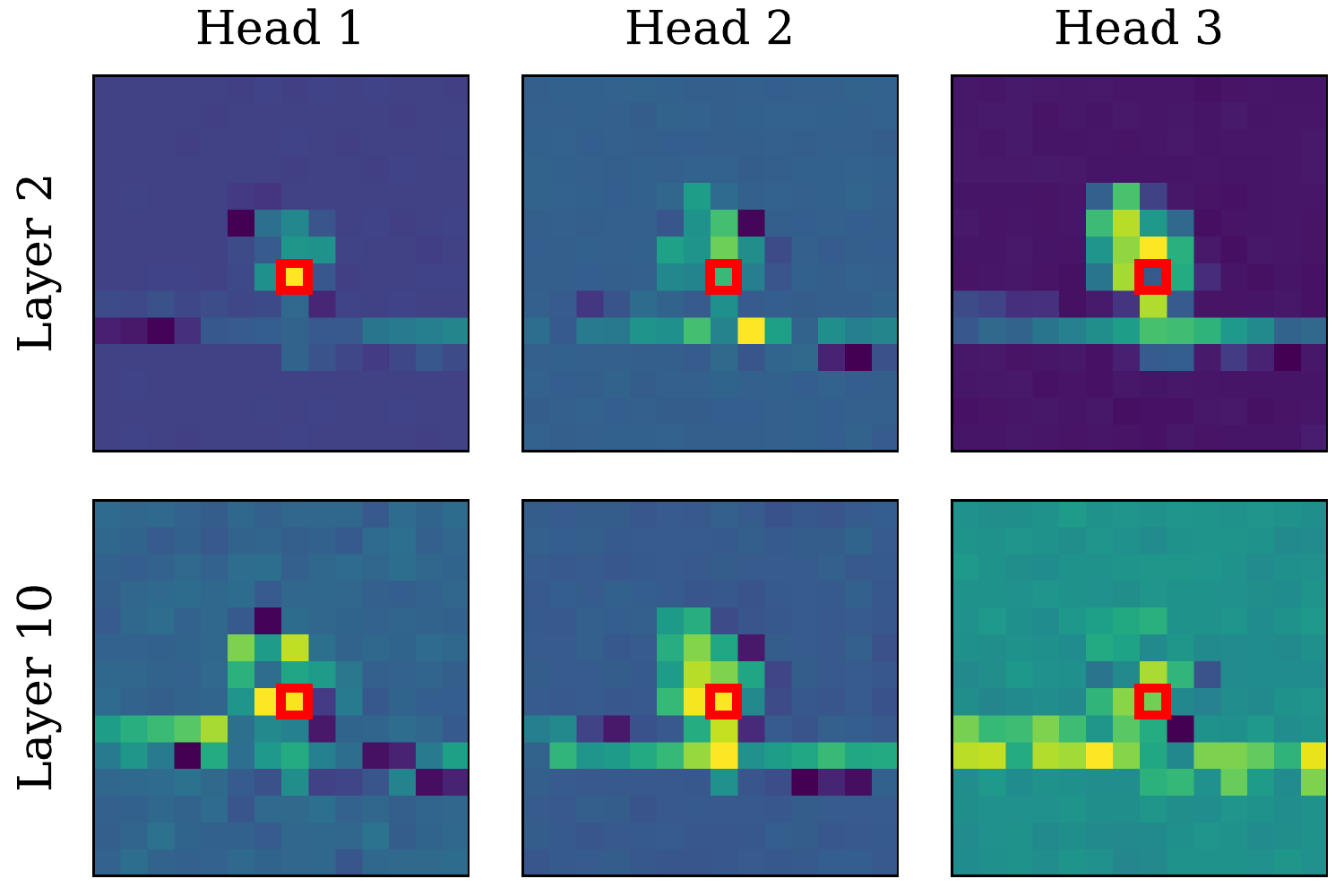}
    \hfill
    \includegraphics[width=.4\linewidth]{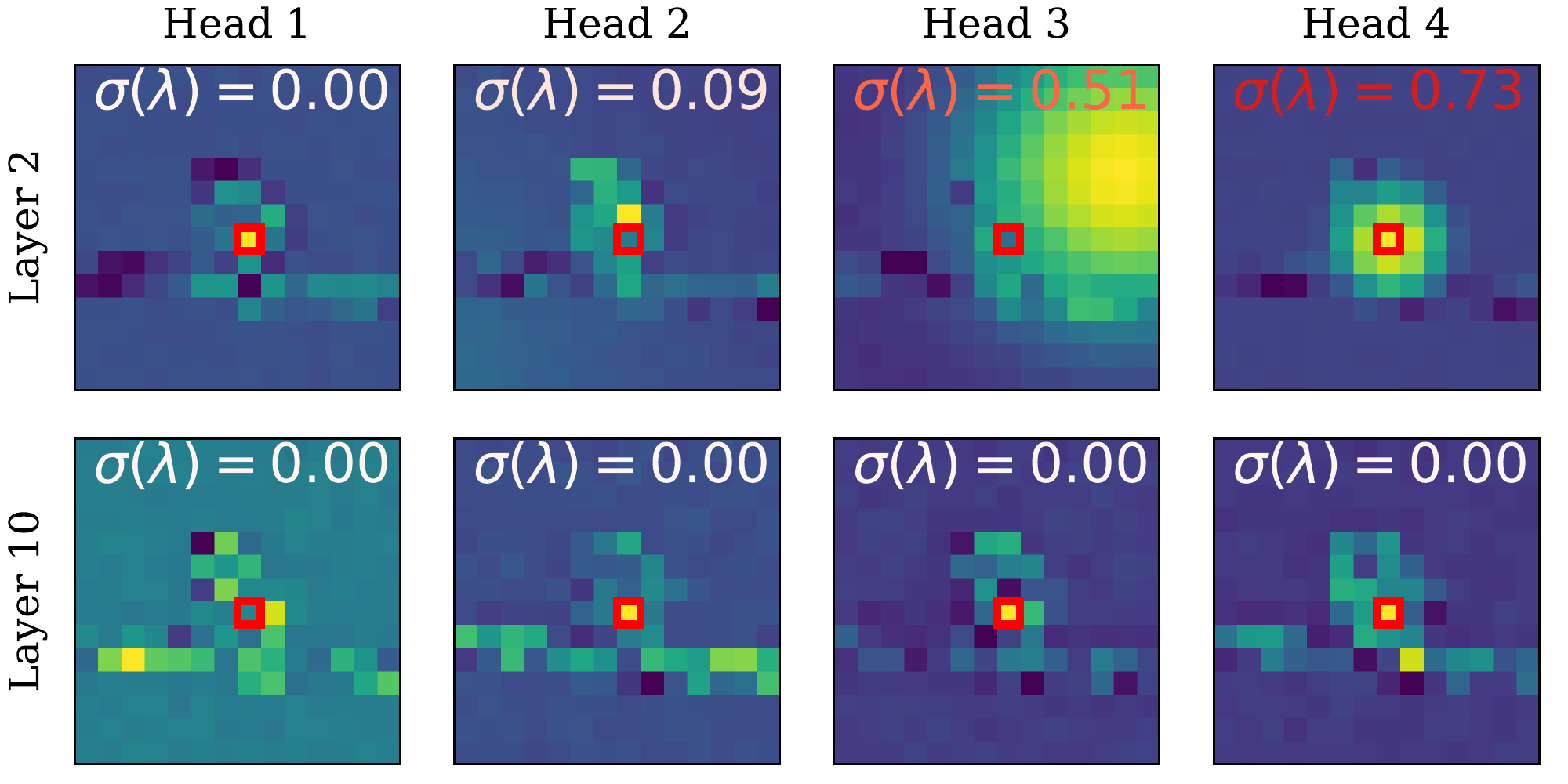}
    \begin{subfigure}[b]{.25\linewidth}
    \includegraphics[width=\linewidth]{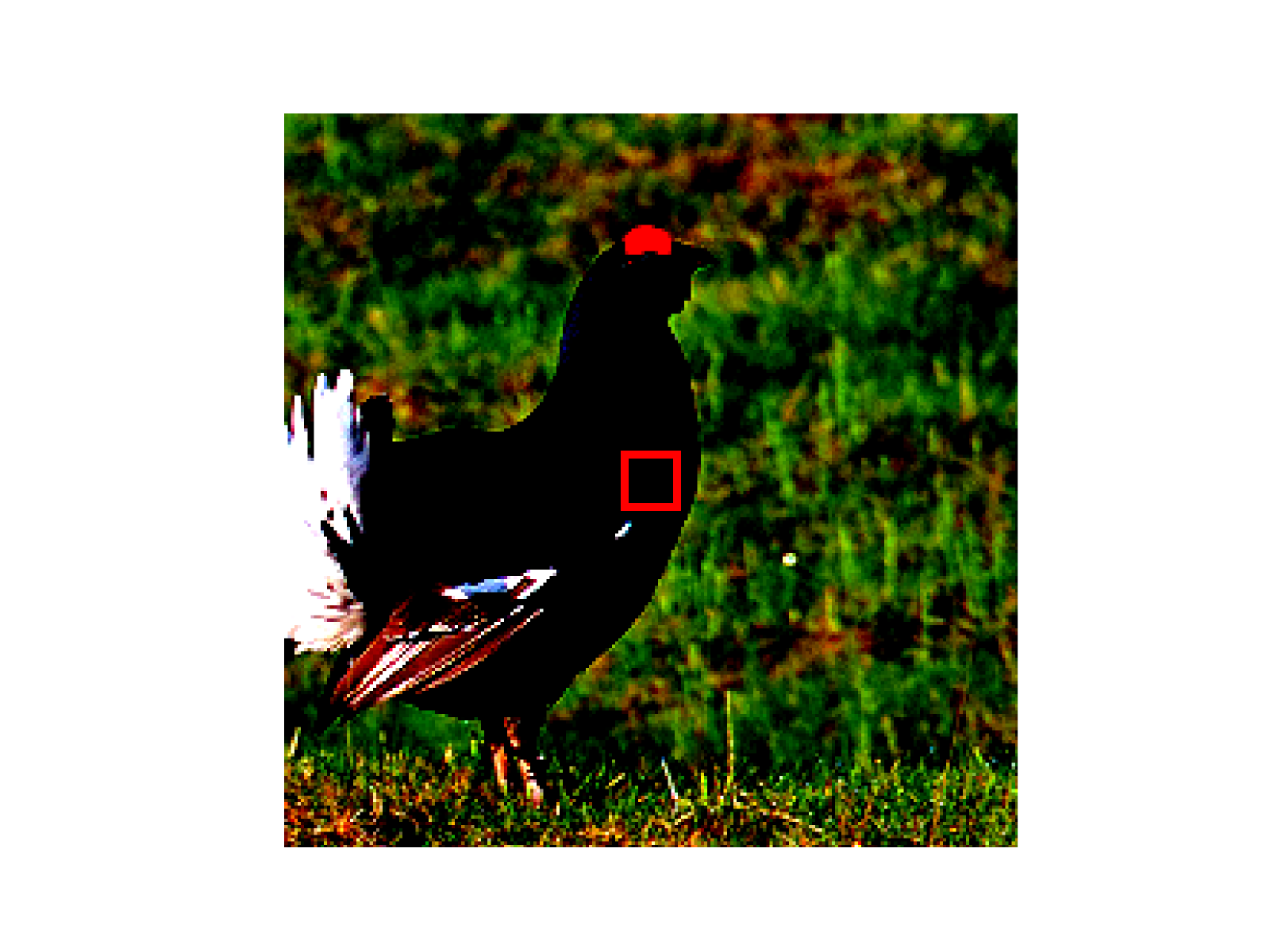}
    \caption{Input images}
    \end{subfigure}
    \begin{subfigure}[b]{.3\linewidth}
    \includegraphics[width=\linewidth]{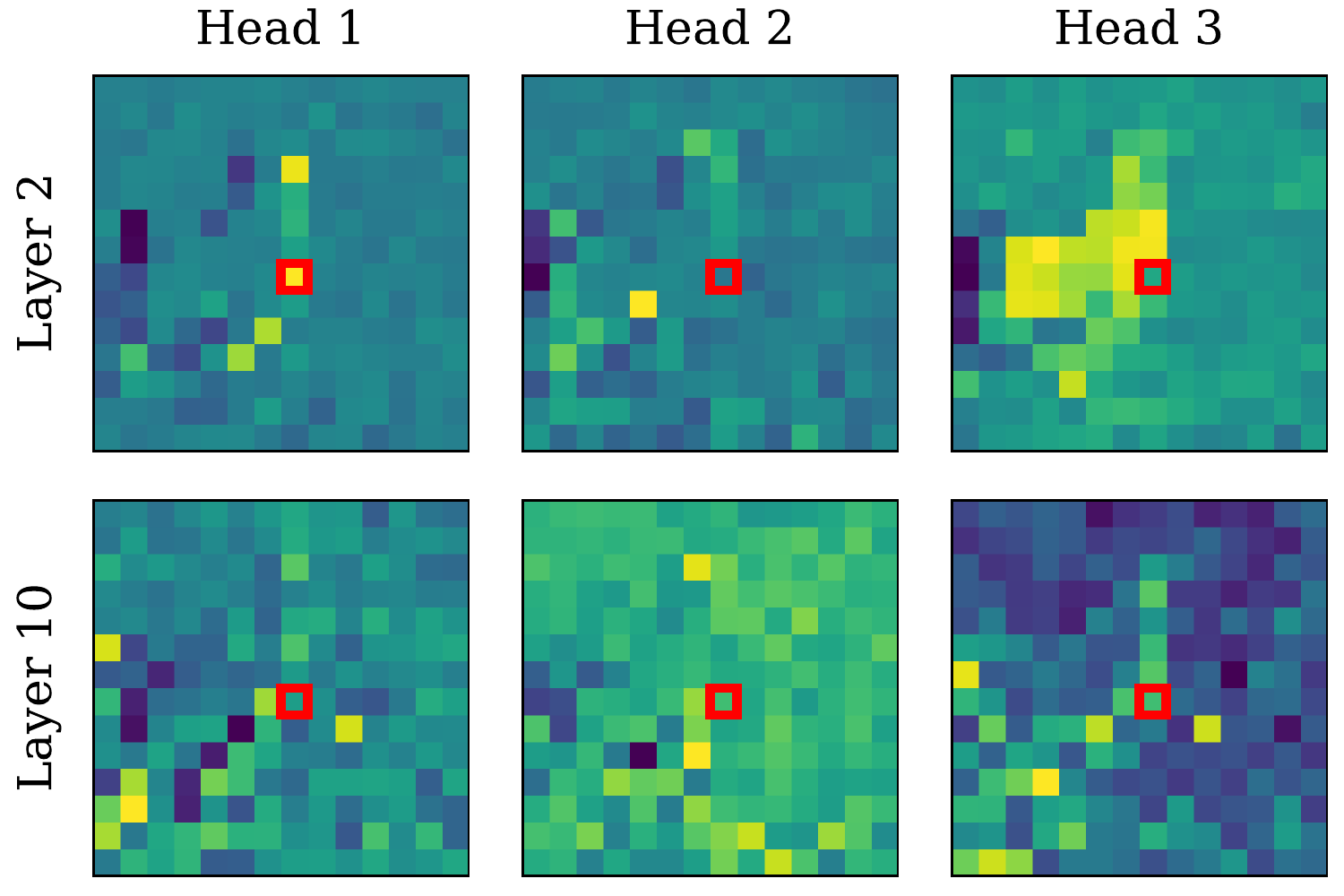}    
    \caption{DeiT}
    \end{subfigure}\hfill
    \begin{subfigure}[b]{.4\linewidth}
    \includegraphics[width=\linewidth]{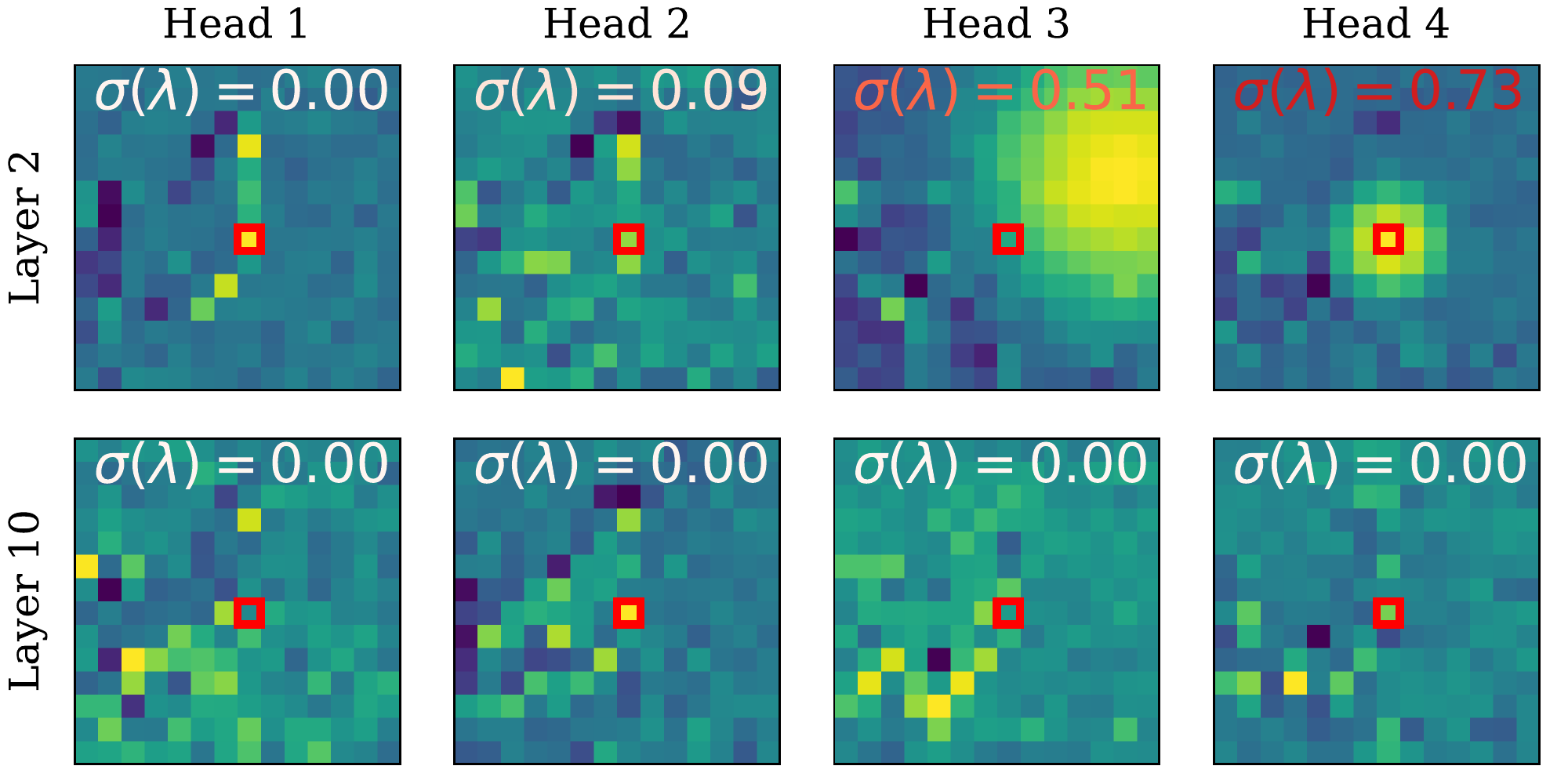}
    \caption{ConViT}
    \end{subfigure}    
    \caption{\textit{Left:} input image which is embedded then fed into the models. The query patch is highlighted by a red box and the colormap is logarithmic to better reveal details. \textit{Center:} attention maps obtained by a DeiT-Ti after 300 epochs of training on ImageNet. \textit{Right:} Same for ConViT-Ti. In each map, we indicated the value of the gating parameter in a color varying from white (for heads paying attention to content) to red (for heads paying attention to position). }
    \label{fig:maps-images}
\end{figure*}

\begin{figure*}[htb]
    \centering
    \begin{subfigure}[b]{.25\linewidth}
    \includegraphics[width=\linewidth]{figs/photo_101.pdf}
    \caption{Input image}
    \end{subfigure}
    \begin{subfigure}[b]{.7\linewidth}
    \includegraphics[width=\linewidth]{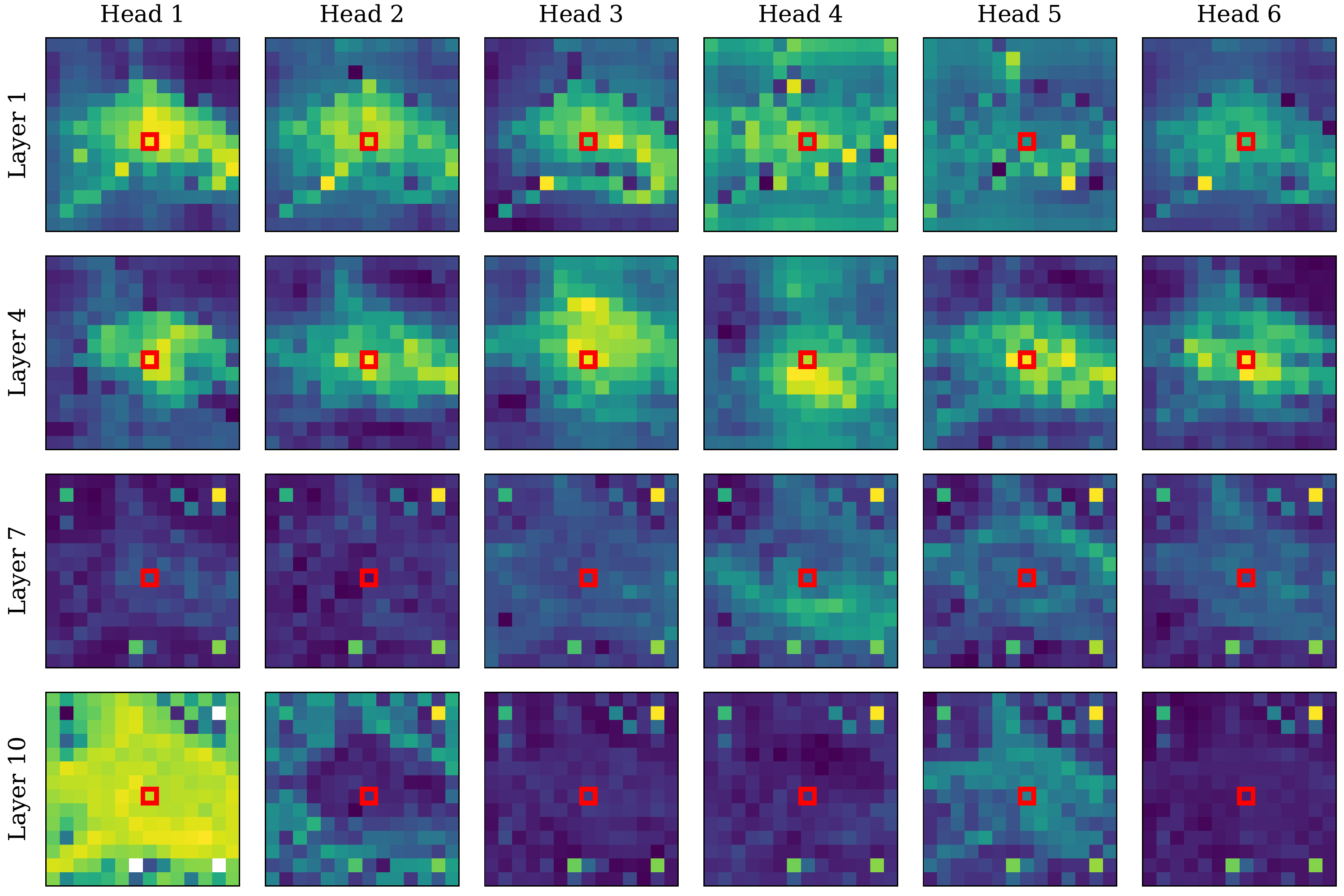}
    \caption{DeiT}
    \end{subfigure}
    \begin{subfigure}[b]{\linewidth}
    \includegraphics[width=\linewidth]{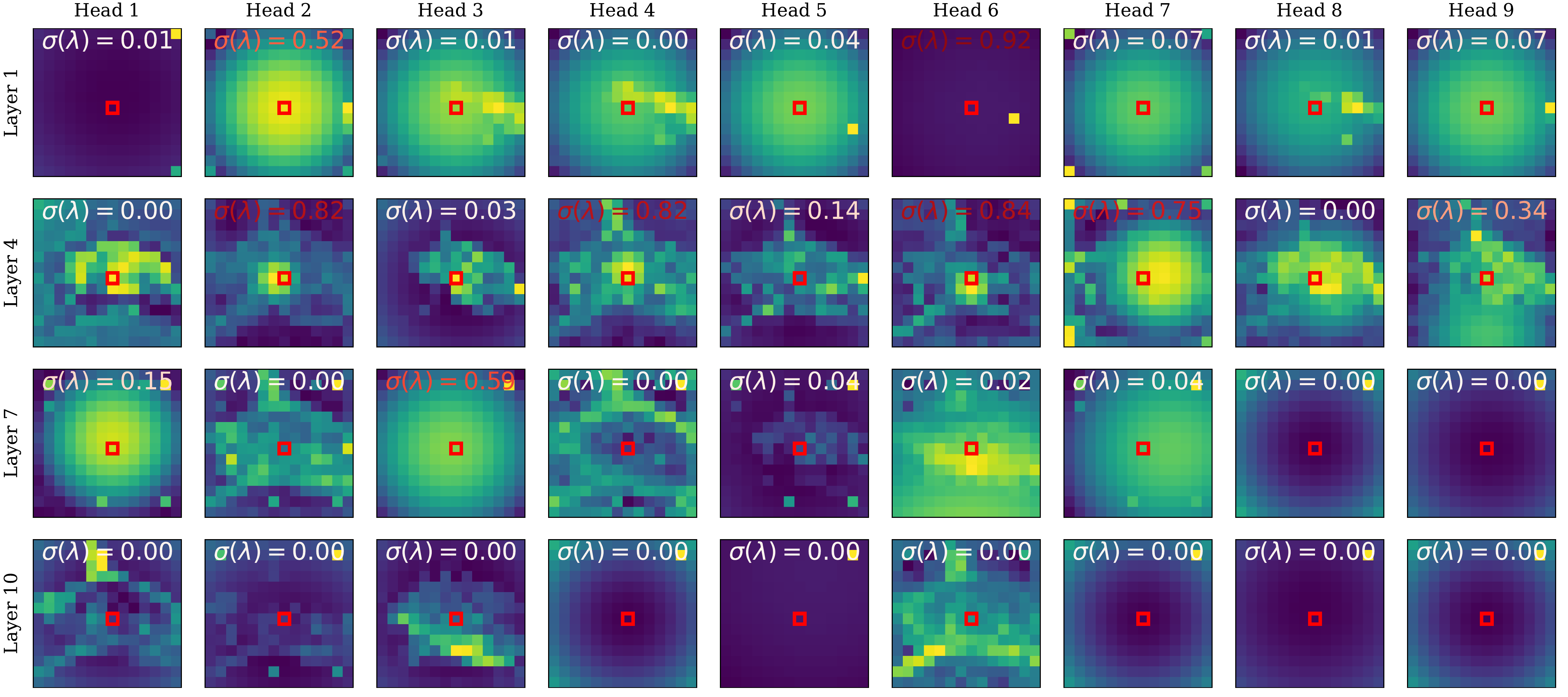}
    \caption{ConViT}
    \end{subfigure}
    \caption{Attention maps obtained by a DeiT-S and ConViT-S after 300 epochs of training on ImageNet. In each map, we indicated the value of the gating parameter in a color varying from white (for heads paying attention to content) to red (for heads paying attention to position). }
    \label{fig:maps-sizes}
\end{figure*}

In Fig.~\ref{fig:maps-images}, we show a comparison of the attention maps of Deit-Ti and ConViT-Ti for different images of the ImageNet validation set. In Fig.~\ref{fig:maps-sizes}, we compare the attention maps of DeiT-S and ConViT-S.

In all cases, results are qualitatively similar: the DeiT attention maps look similar across different heads and different layers, whereas those of the ConViT perform very different operations. Notice that in the second layer, the third and forth head focus stay local whereas the first two heads focus on content. In the last layer, all the heads ignore positional information, focusing only on content.


\clearpage
\section{Further ablations}
\label{app:masking}

In this section, we explore masking off various parts of the network to understand which are most crucial.

In Tab.~\ref{tab:mask-embed}, we explore the importance of the absolute positional embeddings injected to the input in both the DeiT and ConViT. We see that masking them off at test time a mild impact on accuracy for the ConViT, but a significant impact for the DeiT, which is expected as the ConViT already has relative positional information in each of the GPSA layers. This also shows that the absolute positional information contained in the embeddings is not very useful.

In Tab.~\ref{tab:mask-attn}, we explore the relative importance of the positional and content information by masking them off at test time. To do so, we manually set the gating parameter $\sigma(\lambda)$ to 1 (no content attention) or 0 (no positional attention). In the first GPSA layers, both procedures affect performance similarly, signalling that positional and content information are both useful. However in the last GPSA layers, masking the content information kills performance, whereas masking the positional information does not, confirming that content information is more crucial.

\begin{table}[h]
    \footnotesize
    \centering
    \begin{tabular}{c|c|c}
    \toprule
    Model & Mask pos embed & No mask\\
    \midrule
    DeiT-Ti  & 38.3 & 72.2 \\
    ConViT-Ti & 67.1 & 73.1 \\
    \bottomrule
    \end{tabular}
    \caption{Performance on ImageNet with the positional embeddings masked off at test time.}
    \label{tab:mask-embed}
\end{table}

\begin{table}[h]
    \footnotesize
    \centering
    \begin{tabular}{c|c|c|c}
    \toprule
    \# layers masked & Mask content & Mask position & No mask\\
    \midrule
    3 & 62.3 & 63.5 & 73.1 \\
    5 & 35.0 & 53.1 & 73.1 \\
    10 & 1.3 & 46.8 & 73.1 \\
    \bottomrule
    \end{tabular}
    \caption{Performance of ConViT-Ti on ImageNet with positional or content attention masked off at test time.}
    \label{tab:mask-attn}
\end{table}

\clearpage

\end{document}